\documentclass[lettersize,journal]{IEEEtran}
\usepackage{amsmath,amsfonts}
\usepackage{algorithmic}
\usepackage{algorithm}
\usepackage{array}
\usepackage[caption=false,font=normalsize,labelfont=sf,textfont=sf]{subfig}
\usepackage{textcomp}
\usepackage{stfloats}
\usepackage{url}
\usepackage{verbatim}
\usepackage{graphicx}
\usepackage{cite}
\usepackage{graphicx}
\usepackage{amsmath}
\usepackage{amssymb}
\usepackage{booktabs}
\usepackage{multirow}
\usepackage{ulem}
\usepackage{color}
\usepackage{pifont}
\usepackage{balance}
\usepackage{colortbl}
\usepackage{xcolor}
\usepackage{makecell}
\usepackage{amsmath,amssymb,mathrsfs}
\usepackage{bm}
\usepackage{algorithm, algorithmic}

\usepackage{balance}
\usepackage{bbding} 
\usepackage{multirow}
\definecolor{newcolor}{rgb}{.8,.349,.1}

\begin{document}

\title{Towards Comprehensive Monocular Depth Estimation: Multiple Heads Are Better Than One }

\author{Shuwei Shao$^{\dag}$, Ran Li$^{\dag}$, Zhongcai Pei, Zhong Liu, Weihai Chen$^{*}$, \\ Wentao Zhu, Xingming Wu and Baochang Zhang$^{*}$
\thanks{This work was supported by the National Natural Science Foundation of China under grant 61620106012.}
\thanks{Shuwei Shao, Ran Li, Zhongcai Pei, Zhong Liu, Weihai Chen, Xingming Wu and Baochang Zhang are with the School of Automation Science and Electrical Engineering, Beihang University, Beijing, China. (email: swshao@buaa.edu.cn, rnlee1998@buaa.edu.cn, whchen@buaa.edu.cn, bczhang@buaa.edu.cn)}
\thanks{Wentao Zhu is with the Amazon, USA. (email: wentaoz1@uci.edu)}
\thanks{$^{\dag}$ Equal contributions.}
\thanks{*(Joint corresponding author: Weihai Chen and Baochang Zhang.)}
}

\markboth{IEEE Transactions on Multimedia, VOL. XX, NO. XX, XXXX 2022}
{Shao \MakeLowercase{\textit{et al.}}: Towards Comprehensive Monocular Depth Estimation: Multiple Heads Are Better Than One}


\maketitle

\begin{abstract}
Depth estimation attracts widespread attention in the computer vision community. However, it is still quite difficult to recover an accurate depth map using only one RGB image. We observe a phenomenon that existing methods tend to fail in different cases, caused by differences in network architecture, loss function and so on. In this work, we investigate into the phenomenon and propose to integrate the strengths of multiple weak depth predictor to build a comprehensive and accurate depth predictor, which is critical for many real-world applications,~\textit{e.g.}, 3D reconstruction. Specifically, we construct multiple base (weak) depth predictors by utilizing different Transformer-based and convolutional neural network (CNN)-based architectures. Transformer establishes long-range correlation while CNN preserves local information ignored by Transformer due to the spatial inductive bias. Therefore, the coupling of Transformer and CNN contributes to the generation of complementary depth estimates, which are essential to achieve a comprehensive depth predictor. Then, we design mixers to learn from multiple weak predictions and adaptively fuse them into a strong depth estimate. 
The resultant model, which we refer to as Transformer-assisted depth ensembles (TEDepth). On the standard NYU-Depth-v2 and KITTI datasets, we thoroughly explore how the neural ensembles affect the depth estimation and demonstrate that our TEDepth achieves better results than previous state-of-the-art approaches. To validate the generalizability across cameras, we directly apply the models trained on NYU-Depth-v2 to the SUN RGB-D dataset without any fine-tuning, and the superior results emphasize its strong generalizability. 
\end{abstract}

\begin{IEEEkeywords}
Monocular Depth Estimation, Ensemble Learning, Deep Learning, Transformer, Convolutional Neural Network
\end{IEEEkeywords}

\section{Introduction}
\label{sec:intro}
	
\begin{figure}[!htb]
	\centering
	\includegraphics[width=1.0\linewidth]{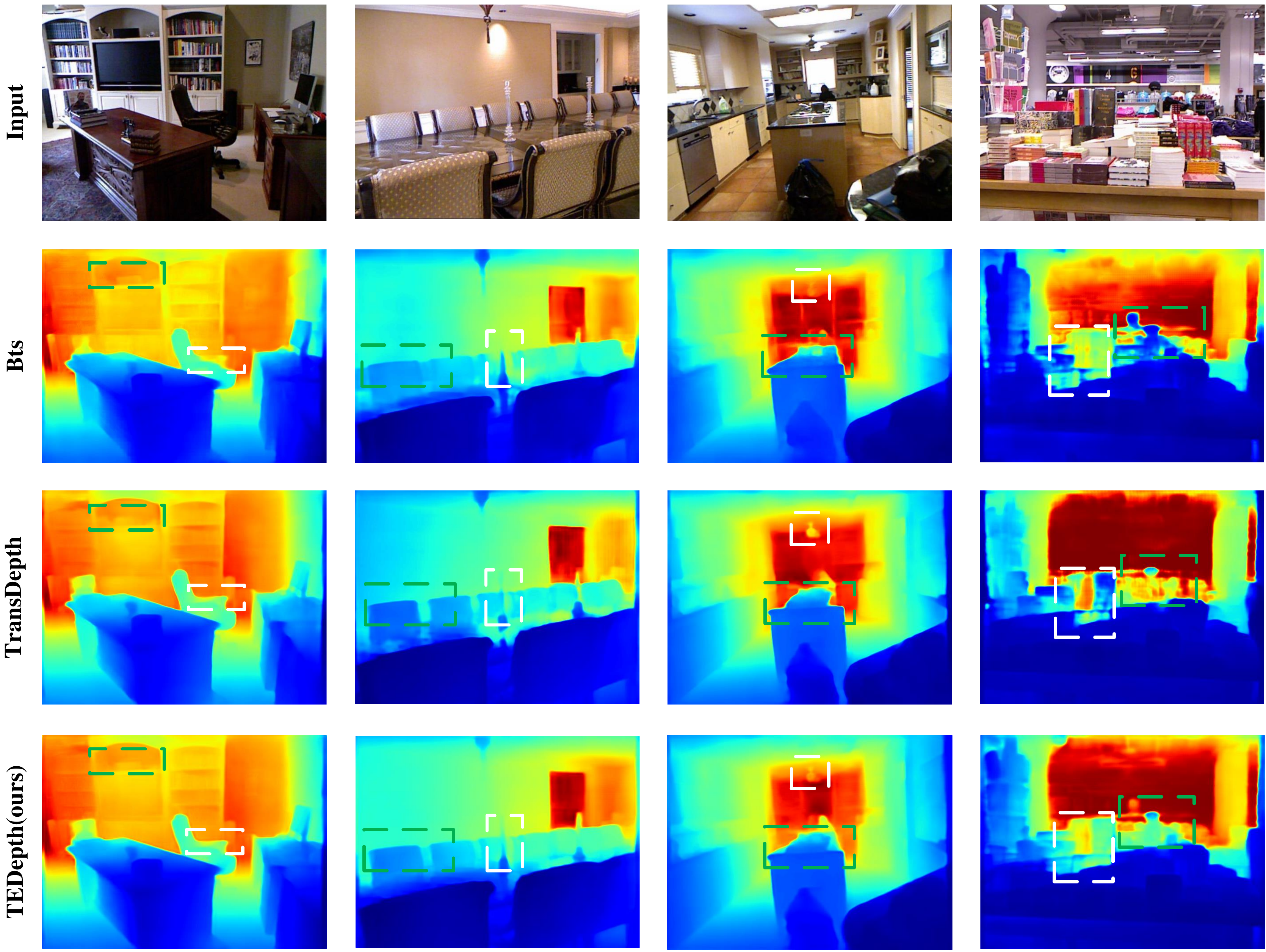}
	\caption{Illustration of the failure cases in BTS~\cite{lee2019big} and TransDepth~\cite{yang2021transformer}, which are highlighted by white and green boxes. By contrast, our TEDepth is capable of achieving more comprehensive and accurate depth estimates.} 
	\label{Fig1}
\end{figure}

Estimating depth map from single RGB image has been a longstanding research topic and proved to be a practical technology, with applications ranging from scene understanding~\cite{hazirbas2016fusenet}, augmented reality~\cite{lee2011depth} through to minimally invasive surgery~\cite{shao2022self}. Saxena~\textit{et al.}~\cite{saxena2005learning} proposed one of the first learning-based studies in this area, and significant advancements have been made followed by the explosion of deep learning~\cite{eigen2015predicting,laina2016deeper,li2017two,fu2018deep,song2019contextualized,yang2019bayesian,ling2021unsupervised}. Nonetheless, their depth estimation performance is still unsatisfactory.


The motivation for this work stems from the asymmetric depth error. To elaborate on it, we present several failure cases of the state-of-the-art monocular depth estimation methods on the NYU-Depth-v2 dataset~\cite{silberman2012indoor} in Fig.~\ref{Fig1}. BTS and TransDepth fail on different regions of the same input image, as can be shown. For example, in the first column, BTS predicts the clear boundary of black box, which however fades in the estimate of TransDepth. BTS predicts the depth of chair armrest wrongly as the background depth, but TransDepth succeeds. To alleviate their failure cases, we intuitively design a depth estimator based on the strengths of each depth prediction, achieving a comprehensive and accurate depth estimation. 

Ensemble learning~\cite{dong2020survey} is an efficacious machine learning paradigm, which combines predictions from individual models to produce superior results. Ensembles of neural networks, or neural ensembles for short, are now playing an essential role in ensemble learning owing to the dominance of deep learning. For instance, neural ensembles have been successfully applied to improve the classification accuracy~\cite{huang2017snapshot,yang2021local} and robustness of object detection~\cite{solovyev2019weighted}. Despite the steady progress in other fields of neural ensemble, the utility and impact of neural ensembles for monocular depth estimation still remains unknown and to be explored.

\begin{figure*}[!htb]
	\centering
	\includegraphics[width=1.0\linewidth]{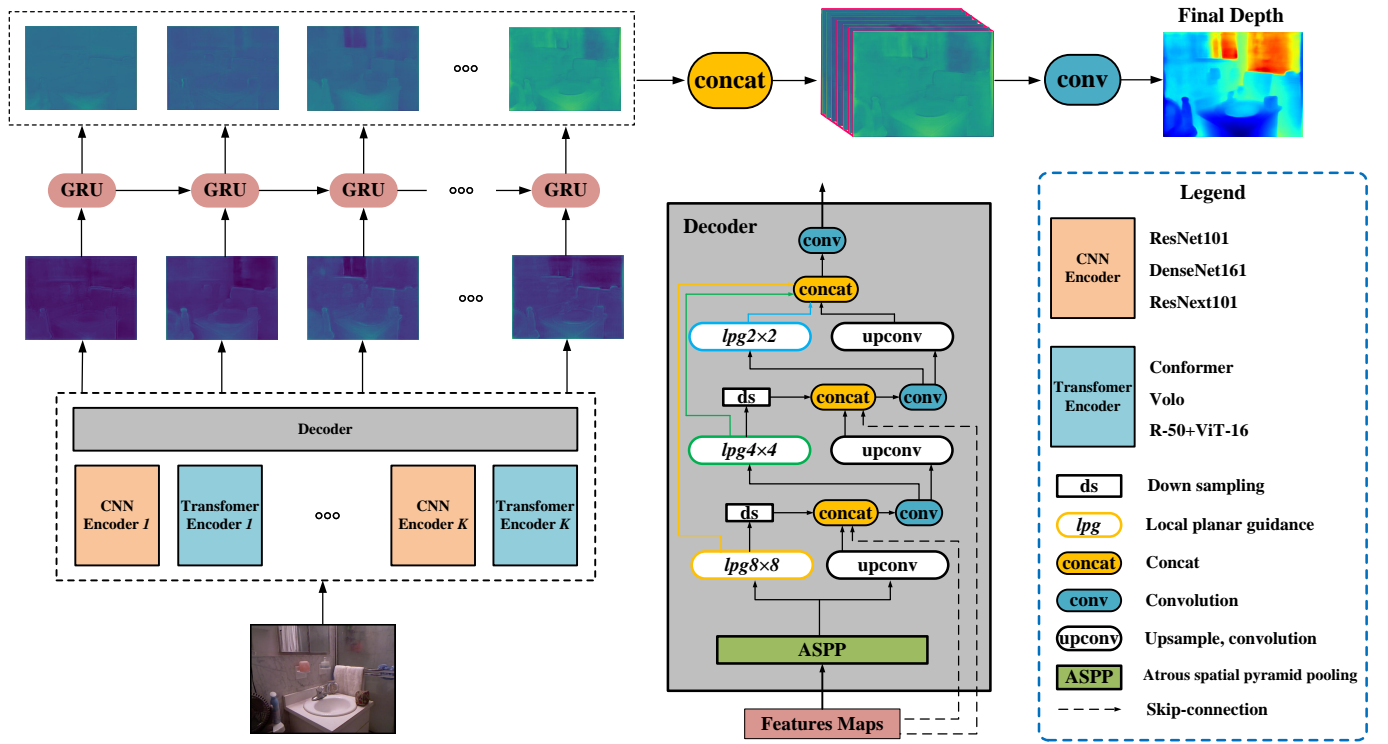}
	\caption{Overview of the proposed TEDepth, which consists of multiple base predictors and a mixer. The weights of decoder are not shared, and the details of local planar guidance and atrous spatial pyramid pooling are in~\cite{lee2019big} and~\cite{chen2017deeplab}, respectively. We illustrate the GRU-based fusion and fuse the feature maps of the penultimate layer instead of the final depth maps, because multi-channel feature maps include more exploitable information than single-channel depth maps.}
	\label{Fig2}
\end{figure*}

We delve into neural ensembles for monocular depth estimation and introduce a two-level ensemble scheme, TEDepth. An overview is demonstrated in Fig.~\ref{Fig2}, where the base (weak) predictors produce asymmetric bottom-level predictions, and the mixer integrates them into a comprehensive depth estimate at the top level. As indicated in recent studies~\cite{lee2019big,ranftl2021vision}, the encoder is more pivotal than the decoder in an encoder-decoder based architecture for monocular depth estimation. Therefore, we focus on the encoder design and use the decoder of BTS for simplicity to construct multiple base predictors. We adopt different Transformer-based architectures~\textit{e.g.},~\cite{yuan2021volo,Peng_2021_ICCV, dosovitskiy2020image} and CNN-based architectures~\textit{e.g.},~\cite{He_2016_CVPR,huang2017densely,xie2017aggregated} as encoders in parallel and prohibit the flow of information among them. It should be noted that the TEDepth is agnostic to the specific selection of network architectures and other choices are feasible. The principle of coupling Transformer and CNN lies in the fact that Transformer establishes the long-range correlation while CNN preserves the local information, which contributes to the generation of complementary depth estimates. In addition, four types of mixers are devised, namely, uniformly weighted fusion, confidence-guided fusion, concatenation-based fusion and ranking-based fusion, to learn from multiple weak predictions and adaptively merge them into a strong depth estimate.


Extensive experiments are conducted on challenging benchmarks including NYU-Depth-v2~\cite{silberman2012indoor}, SUN RGB-D~\cite{song2015sun} and KITTI~\cite{geiger2013vision}, which involve comparison to preceding state-of-the-art depth competitors and ensemble scheme competitors, number of base predictors, mixer types, diversity among base predictors and fusion locations. Besides, we verify its generalizability by applying the models trained on NYU-Depth-v2 to the SUN RGB-D without any finetuning. 

To summarize, the contributions of this work are threefold:

	

\begin{itemize}
	\item Taking inspiration from the asymmetric depth error, we introduce an efficacious framework, termed TEDepth, to achieve a comprehensive and accurate depth predictor by integrating the strengths of weak depth predictors, which provides a novel perspective in model design. 
	\item We propose to combine CNN and Transformer to acquire asymmetric base predictors, which contribute to deliver complementary depth estimates. Besides, we perform an in-depth investigation regarding how the neural ensembles impact the monocular depth estimation. 
	\item We conduct extensive experiments on three challenging datasets, including NYU-Depth-v2, SUN RGB-D and KITTI, demonstrating that our TEDepth outperforms previous methods by a significant margin and achieves a new state-of-the-art.
\end{itemize}

\section{Related Work}

\subsubsection{Monocular depth estimation}
As one of the first learning-based studies, Saxena~\textit{et al.}~\cite{saxena2005learning} introduced a Markov Random Field to regress depth map from single RGB image. After that, Eigen~\textit{et al.}~\cite{eigen2014depth} proposed multi-scale networks and a scale invariant loss. Since then, many follow-up studies have been developed to steadily improve the accuracy. Laina~\textit{et al.}~\cite{laina2016deeper} used a fully convolutional model involving four up-projection modules. Cao~\textit{et al.}~\cite{cao2017estimating} and Fu~\textit{et al.}~\cite{fu2018deep} framed the depth estimation as a classification problem. Qi~\textit{et al.}~\cite{qi2018geonet} built a network named GeoNet to enforce geometrically consistent depth map and surface normal. Lee~\textit{et al.}~\cite{lee2019big} designed local planar guidance layers to recover depth map resolution. Yin~\textit{et al.}~\cite{yin2019enforcing} suggested an idea of incorporating virtual surface normal into the loss calculation, which allowed exploiting geometry information. Yang~\textit{et al.}~\cite{yang2021transformer} leveraged Transformer to capture the long-distance dependencies. 
Vaishakh~\textit{et al.}~\cite{patil2022p3depth} proposed to leverage information from coplanar pixels.

Our work significantly departs from the described methods. Instead, we draw inspiration from the asymmetric depth error and introduce Transformer-assisted depth ensembles to build a comprehensively effective depth estimator.

\subsubsection{Visual transformers} Transformer has received extensive attention as its effectiveness in natural language processing (NLP) tasks~\cite{vaswani2017attention}. Recently, Dosovitskiy~\textit{et al.}~\cite{dosovitskiy2020image} developed the first attempt to indicate the feasibility of Transformer architectures for image classification. Then, the visual transformers were further advanced in~\cite{Peng_2021_ICCV,yuan2021volo,Liu_2021_ICCV}, for example and introduced to more generic tasks, such as semantic segmentation~\cite{zheng2021rethinking}, depth estimation~\cite{yang2021transformer}, weakly supervised object localization~\cite{gao2021ts} and image generation~\cite{jiang2021transgan}. We are inspired by the recent success of Transformer and propose to couple CNN with Transformer to acquire the diverse base predictors in TEDepth.

\subsubsection{Ensemble learning} It can be traced back to the 1990s, and is an efficacious machine learning paradigm that creates a collection of models and utilizes predictions from individual models to perform a superior estimation~\cite{dong2020survey}. Bagging and Boosting~\cite{bauer1999empirical} are one of the seminal works in the early stage of this field. The branch neural ensembles~\cite{ zhou2002ensembling}, particularly ensembles of CNNs~\cite{huang2017snapshot,yang2021local}, hold an essential position in modern ensemble learning. A fundamental problem for neural ensembles is to ensure the diversity among base predictors. The sources of diversity involve using random initializations~\cite{lakshminarayanan2016simple}, different  hyperparameters~\cite{wenzel2020hyperparameter} and additional constraint loss terms~\cite{yang2020dverge}. 

In contrast to the prior ensemble methods, we explore the ensembles of visual transformers and utilize the complementarity of CNN and Transformer to enforce the diversity among base predictors. Despite the substantial research, neural ensembles for monocular depth estimation remain to be explored. The utility and impact are thoroughly evaluated in this work.

\section{Transformer-Assisted Depth Ensembles}
\subsection{Problem setup}

Let $D = \left\{ {\left( {{{\bm{{\rm r}}}_n},{{\bm{{\rm d}}}_n}} \right)} \right\}_1^N$ define a training set containing $N$ pairs of RGB images and ground-truth depth maps. In the previous supervised learning paradigm, a DepthNet parameterized by $\theta$ is trained to learn a mapping function \begin{equation} {f_\theta }:{\bm{{\rm r}}_n} \to {{\hat {\bm{{\rm d}}}}_n}, \end{equation} which converts a RGB image ${\bm{{\rm r}}_n}$ into the corresponding depth map ${\hat {\bm{{\rm d}}}_n}$. Unlike prior works, we propose to (1) learn a collection of asymmetric mapping functions $\left\{ {{f_{{\theta ^i}}}} \right\}_1^K $ through $K$ diverse base predictors, where $K$ denotes the number of base predictors, and (2) leverage the complementary information encoded in each predictions via the mixer. 

\subsection{Base Predictors}

\subsubsection{Network architectures} 
To construct multiple base predictors, we leverage CNN-based architectures ResNet~\cite{He_2016_CVPR}, DenseNet~\cite{huang2017densely} and ResNext~\cite{xie2017aggregated} and Transformer-based architectures  Conformer~\cite{Peng_2021_ICCV}, Volo~\cite{yuan2021volo} and R50+ViT-B/16~\cite{dosovitskiy2020image} as encoders in parallel, and adopt the decoder of BTS~\cite{lee2019big}. The Transformer-based predictors capture the complicated spatial transformation and long-range dependencies that comprise a global representation. Local feature details, on the other hand, are prone to be ignored. The CNN-based predictors are able to collect local features through convolutional operations. Besides, we find that in practice, random initialization of network parameters, random data shuffling and random augmentation, also enforce the diversity among base predictors. 

\subsubsection{Training schedules}  There are roughly three schedules used to train base predictors: independent training, simultaneous training and sequential training~\cite{islam2003constructive}.

In the independent training, each base predictor is trained separately.  Most ensemble models adopt the independent training schedule because of its simplicity and memory savings, in~\cite{russakovsky2015imagenet,liao2018defense}, for example. For the simultaneous training, the parameters of all base predictors are updated together in each training iteration. The simultaneous training schedule allows involving interaction such as mutual learning~\cite{zhang2018deep} of base predictors in the ensemble. However, this comes at a cost of huge memory consumption. The sequential training schedule can be viewed as a compromise between the independent and simultaneous training schedules, where the base predictors are trained sequentially. In other words, the parameters of the trained base predictors are frozen before training the next base predictor, enabling unidirectional interaction. 

The training schedules are primarily explored on the image classification and may not be entirely suitable for depth estimation because its network typically consists of a complicated encoder-decoder architecture. In addition, the numerous matrix multiplication operations in the Transformer determine the massive memory demand. Hence, we employ the independent training schedule. Since the base predictors do not interact with one another, they are more likely to fall into distinct local optimums. Inconsistencies in the base predictors contribute to the generation of asymmetric predictions. 

\begin{algorithm}
	\renewcommand{\algorithmicrequire}{\textbf{Input:}}
	\renewcommand{\algorithmicensure}{\textbf{Output:}}
	\caption{The TEDepth training procedure}
	\label{alg1}
	\begin{algorithmic}[1]
		\REQUIRE Training set $D = \left\{ {\left( {{{\bm{{\rm r}}}_n},{{\bm{{\rm d}}}_n}} \right)} \right\}_1^N$; randomly split $D$ into ${D_{train\_base}}$ and ${D_{train\_mixer}}$ at a 7:1 ratio;
		
		Level-0 base predictors: $\left\{ {{f_{{\theta ^i}}}} \right\}_1^K$;
		
		Level-1 mixer: ${M_\theta }$.
		
		\ENSURE Trained base predictors $\left\{ {{f_{{\theta ^i}}}} \right\}_1^K$ and mixer ${M_\theta }$.
		
		\textbf{/*} Train level-0 base predictors on the ${D_{train\_base}}$. \textbf{*/}
		
		\STATE \textbf{Initialization:} Randomly initialize $\left\{ {{{{\theta ^i}}}} \right\}_1^K$;
		
		\FOR{$i = 1, \ldots ,K$}
		\STATE Compute depth predictions of base predictor ${f_{{\theta ^i}}}$; 
		\STATE Calculate the loss function in Eq.~\ref{eq3}; 
		\STATE Update the model parameters $\left\{ {{{{\theta ^i}}}} \right\}$ by AdamW~\cite{loshchilov2017decoupled}.	
		\ENDFOR			
		
		\textbf{/*} Train level-1 mixer on the ${D_{train\_mixer}}$. \textbf{*/} 
		\STATE \textbf{Initialization:} Randomly initialize $\theta$;
		\STATE Freeze network parameters of base predictors $\left\{ {{f_{{\theta ^i}}}} \right\}_1^K$; 
		\STATE Compute feature predictions of $\left\{ {{f_{{\theta ^i}}}} \right\}_1^K$; 
		\STATE Compute depth predictions of mixer ${M_\theta }$; 
		\STATE Calculate the loss function in Eq.~\ref{eq3}; 
		\STATE Update the model parameters $\theta$ by AdamW.	
		
	\end{algorithmic}  
\end{algorithm}

To avoid biasing the estimation performance, base predictors are only trained on a subset of the whole training set. Concretely, we divide the completed training set into ${D_{train\_base}}$ and ${D_{train\_mixer}}$ at a 7:1 (empirical setting) ratio, and the base predictors are trained on ${D_{train\_base}}$. The predictions on ${D_{train\_mixer}}$ made by base predictors can then be used to train the mixer. Algorithm~\ref{alg1} summarizes the overall training procedure.

\subsubsection{Loss function} We use a scaled scale-invariant (SSI) loss introduced by Lee~\textit{et al.}~\cite{lee2019big}, 

\begin{equation}
	{{\mathcal L}_{SSI}}=\alpha \sqrt{\frac{1}{\left| \text{T} \right|}\sum\limits_{j}{{{\left( {{\text{g}}^{j}} \right)}^{2}}-\frac{\eta }{{{\left| \text{T} \right|}^{2}}}{{\left( \sum\limits_{j}{{{\text{g}}^{j}}} \right)}^{2}}}}, \label{eq3}
\end{equation} 
where ${{\bm{{\rm g}}}^j} = \log {{\widehat {\bm{{\rm d}}}}^j} -   \log {{\bm{{\rm d}}}^j}$, ${{\bm{{\rm d}}}}$ is the ground-truth depth map, and ${{\bm{{\rm T}}}}$ stands for a set of pixels with valid ground-truth values. The $\alpha$ and $\eta $ are set to 10 and 0.85 based on~\cite{lee2019big}.

\subsection{Ensemble fusion}
The mixer ${M_\theta }$ is leveraged for integrating the complementary predictions from base predictors. Formally, 
\begin{equation} 
	{\widehat {\bm{{\rm d}}}} = {M_\theta }\left( {\left\{ {{f_{{\theta ^i}}}\left( {{{\bm{{\rm r}}}}} \right)} \right\}_1^K} \right).
\end{equation}

We did not fuse the final depth maps, but rather the feature maps of the penultimate layer, because multi-channel feature maps contain more exploitable information than single-channel depth maps.

The mixer contains four different types, uniformly weighted fusion, confidence-guided fusion, concatenation-based fusion and ranking-based fusion. After that, a convolutional layer can be added to generate the final depth estimate.

\subsubsection{Uniformly weighted fusion} The predictions from base predictors are combined with uniform weight,
\begin{equation}
	{{\bm{{\rm F}}}} = \sum\limits_{i = 1}^K {{f_{{\theta ^i}}}\left( {{{\bm{{\rm r}}}}} \right)}, 
\end{equation}
where ${{\bm{{\rm F}}}}$ denotes the fused feature map.

\subsubsection{Confidence-guided fusion} The predictions are integrated using confidence maps,
\begin{equation}
	{{\bm{{\rm F}}}} = \sum\limits_{i = 1}^K {{{\bm{{\rm C}}}^i}\odot{f_{{\theta ^i}}}\left( {{{\bm{{\rm r}}}}} \right)},
\end{equation}
with \begin{equation}{{\bm{{\rm C}}}^i} = {\sigma _1}\left( {Con{v_{3 \times 3}}\left( {{f_{{\theta ^i}}}\left( {{{\bm{{\rm r}}}}} \right)} \right)} \right), \end{equation}
where ${{\bm{{\rm C}}}}$ stands for the confidence map, $\odot$ stands for the element-wise multiplication and ${\sigma _1}\left(  \cdot  \right)$ denotes the sigmoid activation.

\subsubsection{Concatenation-based fusion} The predictions are fused by the concatenation operation and convolutional layer,
\begin{equation}
	{{\bm{{\rm F}}}} = {\sigma _2}\left( {Con{v_{3 \times 3}}\left( {\left[ {\left\{ {{f_{{\theta ^i}}}\left( {{{\bm{{\rm r}}}}} \right)} \right\}_1^K} \right]} \right)} \right),
\end{equation}
where $\left[  \cdot  \right]$ stands for the concatenation operation and ${\sigma _2}\left(  \cdot  \right)$ is the ELU activation~\cite{clevert2015fast}.

\subsubsection{Ranking-based fusion} The predictions are integrated via a gated activation unit based on the GRU network, with convolutions in place of fully connected layers, 
\begin{equation}
	{{\bm{{\rm z}}}^i} = \sigma _1 \left( {Con{v_{3 \times 3}}\left( {\left[ {{{\bm{{\rm h}}}^{i - 1}},{f_{{\theta ^i}}}\left( {{{\bm{{\rm r}}}_n}} \right)} \right]} \right)} \right),
\end{equation}
\begin{equation}
	{{\bm{{\rm s}}}^i} = \sigma _1 \left( {Con{v_{3 \times 3}}\left( {\left[ {{{\bm{{\rm h}}}^{i - 1}},{f_{{\theta ^i}}}\left( {{{\bm{{\rm r}}}_n}} \right)} \right]} \right)} \right),
\end{equation}
\begin{equation}
	{\widetilde {\bm{{\rm h}}}^i} = \tanh \left( {Con{v_{3 \times 3}}\left( {\left[ {{{\bm{{\rm s}}}^i} \odot {{\bm{{\rm h}}}^{i - 1}},{f_{{\theta ^i}}}\left( {{{\bm{{\rm r}}}_n}} \right)} \right]} \right)} \right),
\end{equation}
\begin{equation}
	{{\bm{{\rm h}}}^i} = \left( {1 - {{\bm{{\rm z}}}^i}} \right) \odot {{\bm{{\rm h}}}^{i - 1}} + {{\bm{{\rm z}}}^i} \odot {\widetilde {\bm{{\rm h}}}^i},
\end{equation}

\begin{equation}
	{{\bm{{\rm F}}}} = {\left[ {\left\{ {{{\bm{{\rm h}}}^i}} \right\}_1^K} \right]},
\end{equation}
where ${{\bm{{\rm z}}}}$ denotes the update gate, ${{\bm{{\rm s}}}}$ denotes the reset gate, and ${{\bm{{\rm h}}}}$ denotes the hidden state. With the deepening of fusion, the accuracy of hidden state in ConvGRU gradually increases. At this point, feeding lower-accuracy feature maps into the ConvGRU may deteriorate the hidden state. 
Hence, the ConvGRU receives feature maps in the order of accuracy from low to high. 

Finally, we use a layer of convolution to convert the fused feature map into the final comprehensive depth map,
\begin{equation}\widehat {\bm{{\rm d}}} = \kappa {\sigma _1}\left( {Con{v_{3 \times 3}}\left( {\bm{{\rm F}}} \right)} \right), \end{equation}
where $\kappa$ denotes the scale factor, and is set to 10 for the NYU-Depth-v2 and SUN RGB-D datasets, and 80 for the KITTI dataset, following the settings in~\cite{lee2019big}.

\begin{table*}[htb!]
	\begin{center}
		\renewcommand{\arraystretch}{1.3}
		\resizebox{1.83\columnwidth}{!}{\begin{tabular}{c || c || c c c || c c c}	
				\Xhline{1.2pt}
				Method &  Cap & Abs Rel $\downarrow$ & RMSE $\downarrow$ & ${\textbf{\rm{log}}_{\bm{{10}}}}$ $\downarrow$ &  $\delta  < 1.25$ $\uparrow$ &  $\delta  < {1.25^2}$ $\uparrow$& $\delta  < {1.25^3}$ $\uparrow$ \\ 
				\hline						
				\hline
				Eigen~\textit{et al.}~\cite{eigen2014depth}& 0-10m& 0.158&0.641&-&0.769&0.950&0.988
				\\   
				Fu~\textit{et al.}~\cite{fu2018deep}& 0-10m&0.115&0.509&0.051&0.828&0.965&0.992
				\\ 
				Qi et al.~\cite{qi2018geonet}&0-10m&0.128&0.569&0.057&0.834&0.960&0.990
				\\ 
				VNL~\cite{yin2019enforcing}& 0-10m&0.108&0.416&0.048&0.875&0.976&0.994
				\\
				BTS~\cite{lee2019big}& 0-10m&0.113&0.407&0.049&0.871&0.977&0.995
				\\
				Chen~\textit{et al.}~\cite{chen2020laplacian}& 0-10m&0.111&0.514&0.048&0.878&0.977&0.994
				\\    
				Zhang~\textit{et al.}~\cite{zhang2020densely}& 0-10m&0.112&0.447&0.048&0.881&0.979&0.996
				\\
				DAV~\cite{huynh2020guiding}& 0-10m&0.108&0.412&-&0.882&0.980&0.996
				\\ 
				Long~\textit{et al.}~\cite{Long_2021_ICCV}& 0-10m&0.101&0.377&0.044&0.890&0.982&0.996
				\\ 
				TransDepth~\cite{yang2021transformer}& 0-10m&0.106&0.365&0.045&0.900&0.983&0.996
				\\ 
				P3Depth~\cite{patil2022p3depth}& 0-10m&0.104&0.356&\textbf{0.043}&0.898&0.981&0.996
				\\ 					
				\hline
				\textbf{TEDepth (ours)} & 0-10m&\textbf{0.100} &\textbf{0.349}&\textbf{0.043}&\textbf{0.907}&\textbf{0.987}&\textbf{0.998}\\
				\Xhline{1.2pt}
		\end{tabular}}
	\end{center}
	\caption{Quantitative depth comparison against previous state-of-the-art approaches on the NYU-Depth-V2 dataset. ``-'' means not applicable. The best results are indicated in bold.  }
	\label{table1}
\end{table*}

\begin{table*}[htb!]
	\begin{center}
		\renewcommand{\arraystretch}{1.3}
		\resizebox{2.0\columnwidth}{!}{\begin{tabular}{c|| c || c c  c c || c c c }	
				\Xhline{1.2pt}
				Method & Cap & Abs Rel $\downarrow$ & Sq Rel $\downarrow$ & RMSE $\downarrow$ & RMSE log $\downarrow$ & $\delta  < 1.25$ $\uparrow$ & $\delta  < {1.25^2}$ $\uparrow$& $\delta  < {1.25^3}$ $\uparrow$ \\ 
				\hline						
				\hline
				Eigen~\textit{et al.}~\cite{eigen2015predicting}&0-80m&0.203&1.548&6.307&0.282&0.702&0.898&0.967
				\\  
				Fu~\textit{et al.}~\cite{fu2018deep}&0-80m&0.072&0.307&2.727&0.120&0.932&0.984&0.994
				\\ 
				VNL~\cite{yin2019enforcing}&0-80m&0.072&-&3.258&0.117&0.938&0.990&0.998
				\\
				BTS~\cite{lee2019big}&0-80m&0.061&0.261&2.834&0.099&0.954&0.992&0.998
				\\ 
				Chen~\textit{et al.}~\cite{chen2020laplacian}&0-80m&0.090&0.546&3.802&0.151&0.902&0.972&0.990
				\\ 	 
				Zhang~\textit{et al.}~\cite{zhang2020densely}&0-80m&0.064&0.265&3.084&0.106&0.952&0.993&0.998
				\\
				TransDepth~\cite{yang2021transformer}&0-80m&0.064&0.252&2.755&0.098&0.956&0.994&\textbf{0.999}
				\\ 
				P3Depth~\cite{patil2022p3depth}&0-80m&0.071&0.270&2.842&0.103&0.953&0.993&0.998
				\\					
				\hline
				\textbf{TEDepth (ours)} &0-80m &\textbf{0.056}&\textbf{0.174}&\textbf{2.223}&\textbf{0.084}&\textbf{0.968}&\textbf{0.996}&\textbf{0.999}\\
				\hline
				\hline
				Fu~\textit{et al.}~\cite{fu2018deep}&0-50m&0.071&0.268&2.271&0.116&0.936&0.985&0.995
				\\
				BTS~\cite{lee2019big}&0-50m&0.058&0.183&1.995&0.090&0.962&0.994&\textbf{0.999}
				\\
				Chen~\textit{et al.}~\cite{chen2020laplacian}&0-50m&0.087&0.440&2.907&0.143&0.913&0.976&0.991
				\\ 
				Zhang~\textit{et al.}~\cite{zhang2020densely}&0-50m&0.061&0.200&2.283&0.099&0.960&0.995&\textbf{0.999}
				\\
				TransDepth~\cite{yang2021transformer}&0-50m&0.061&0.185&1.992&0.091&0.963&0.995&\textbf{0.999}
				\\
				\hline
				\textbf{TEDepth (ours)} &0-50m&\textbf{0.054}&\textbf{0.135}&\textbf{1.664}&\textbf{0.080}&\textbf{0.972}&\textbf{0.997} &\textbf{0.999}\\
				\Xhline{1.2pt}
				
		\end{tabular}}
	\end{center}
	\caption{Quantitative depth comparison on the KITTI dataset. The best results are indicated in bold.}
	\label{table2}
\end{table*}

\section{Experiment}
We conduct an extensive set of experiments on three standard benchmarks for indoor and outdoor scenarios, including NYU-Depth-v2, KITTI and SUN RGB-D. In the following, we first present a description of the relevant datasets, evaluation metrics and implementation details. Then, we provide quantitative and qualitative comparison results to previous state-of-the-art competitors. Finally, we demonstrate generalization and ablation studies to discuss a detailed analysis of TEDepth.

\subsection{Datasets}
\textbf{NYU-Depth-v2} dataset~\cite{silberman2012indoor} contains 120K RGB and depth samples from 464 indoor scenes, which are captured at a resolution of $640 \times 480$ pixels. We adopt the official split as previous works and the dataset processed by Lee~\textit{et al.}~\cite{lee2019big}, resulting in 24231 and 654 image-depth pairs for training and testing, respectively.

\textbf{KITTI} is an outdoor dataset~\cite{geiger2013vision} acquired from 61 scenes using 
equipment placed on a moving vehicle. The resolution of RGB images is around $1241 \times 376$ pixels. To compare with existing works, we follow the commonly used Eigen split~\cite{eigen2014depth}, which includes 29 scenes of 697 images for the test set, and 23488 images covering 32 scenes for the training set.

\textbf{SUN RGB-D} is an indoor dataset~\cite{song2015sun} with roughly 10K images captured with four sensors and a wide range of scene diversity. To validate the generalization ability of models, we randomly collect 500 images.
\subsection{Evaluation metrics}
In line with previous works, we use the standard evaluation metrics in our experiments: 
\begin{itemize}
	\item Abs Rel: $\frac{1}{\left| {\bm{{\rm T}}} \right|}\sum\nolimits_{\widehat {\bm{{\rm d}}} \in {\bm{{\rm T}}}} {\left| {\widehat {\bm{{\rm d}}} - {\bm{{\rm d}}}} \right|} /{\bm{{\rm d}}}$; 
	\item Sq Rel: $\frac{1}{\left| {\bm{{\rm T}}} \right|}{\sum\nolimits_{\widehat {\bm{{\rm d}}} \in {\bm{{\rm T}}}} {\left\| {\widehat {\bm{{\rm d}}} - {\bm{{\rm d}}}} \right\|} ^2}/{\bm{{\rm d}}}$;
	\item RMSE: $\sqrt {\frac{1}{\left| {\bm{{\rm T}}} \right|}{{\sum\nolimits_{\widehat {\bm{{\rm d}}} \in {\bm{{\rm T}}}} {\left\| {\widehat {\bm{{\rm d}}} - {\bm{{\rm d}}}} \right\|} }^2}}$;
	\item ${\log _{10}}$: $\frac{1}{\left| {\bm{{\rm T}}} \right|}{\sum\nolimits_{\widehat {\bm{{\rm d}}} \in {\bm{{\rm T}}}} {\left\| {{{\log }_{10}}\widehat {\bm{{\rm d}}} - {{\log }_{10}}{\bm{{\rm d}}}} \right\|} ^2}$;
	\item $\delta < t$: $\%$ of ${{\bm{{\rm d}}}}$ satisfies $\left( {\max \left( {\frac{{{{\widehat {\bm{{\rm d}}}}}}}{{{{\bm{{\rm d}}}}}},\frac{{{{\bm{{\rm d}}}}}}{{{{\widehat {\bm{{\rm d}}}}}}}} \right) = \delta  < t} \right)$ for $t = 1.25,{1.25^2},{1.25^3}.$ 
\end{itemize}

\subsection{Implementation Details}

\begin{figure*}[!]
	\centering
	\includegraphics[width=1.0\linewidth]{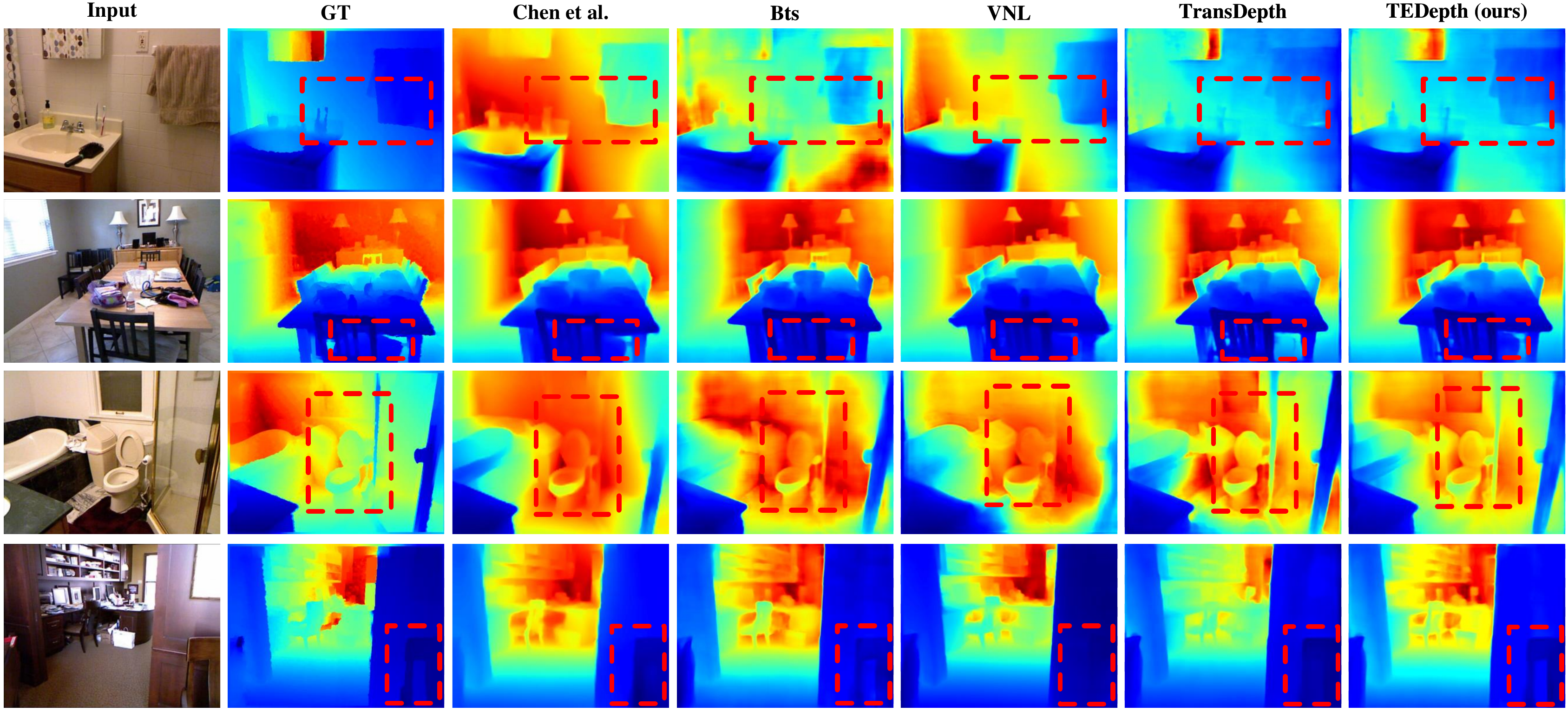}
	\caption{Qualitative depth comparison with previous state-of-the-art approaches on the NYU-Depth-v2 dataset. The red boxes show the regions to focus on. Our TEDepth can derive depth maps with finer-grained details and more clear object boundaries.}
	\label{Fig3}
\end{figure*}

\begin{figure*}[!]
	\centering
	\includegraphics[width=1.0\linewidth]{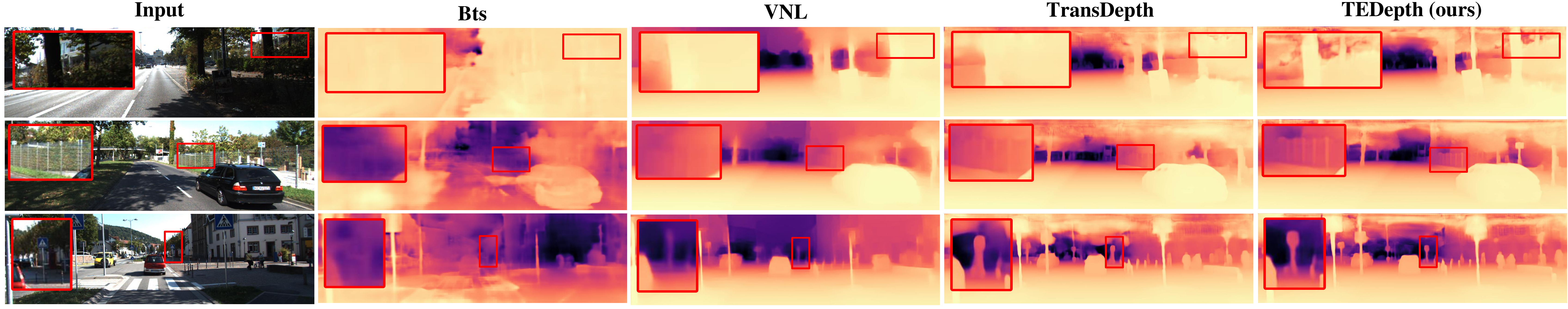}
	\caption{Qualitative depth comparison on the KITTI dataset. The red boxes indicate the areas to emphasize. Our TEDepth demonstrates a better performance on thinner structures signs and posts, for example and performs better to delineate difficult object boundaries.}
	\label{Fig4}
\end{figure*}

The TEDepth is implemented in the PyTorch library~\cite{paszke2017automatic} and trained on four NVIDIA RTX A5000 GPUs. We use the AdamW optimizer~\cite{loshchilov2017decoupled} where ${\beta _1} =$ 0.9, ${\beta _2} =$ 0.999 and $\epsilon=$ 1e-6, and a batch size of 4. The learning rate is scheduled via polynomial decay from a base value of 1e-4 with power $p=$ 0.9. The total number of epochs is set to 50.

To further increase the diversity among base predictors, we perform random horizontal flips and rotations in the ranges [-1, 1] and [-2.5, 2.5] for KITTI and NYU-Depth-v2 datasets, respectively, as well as the following augmentations at 50\% chance: random contrast, brightness, and color adjustment with ranges of $\pm$ 0.1. Based on the random crop, the resolutions of the input are 352 $\times$ 704 pixels for KITTI and 416 $\times$ 544 pixels for NYU-Depth-v2.
\begin{figure*}[!]
	\centering
	\includegraphics[width=0.98\linewidth]{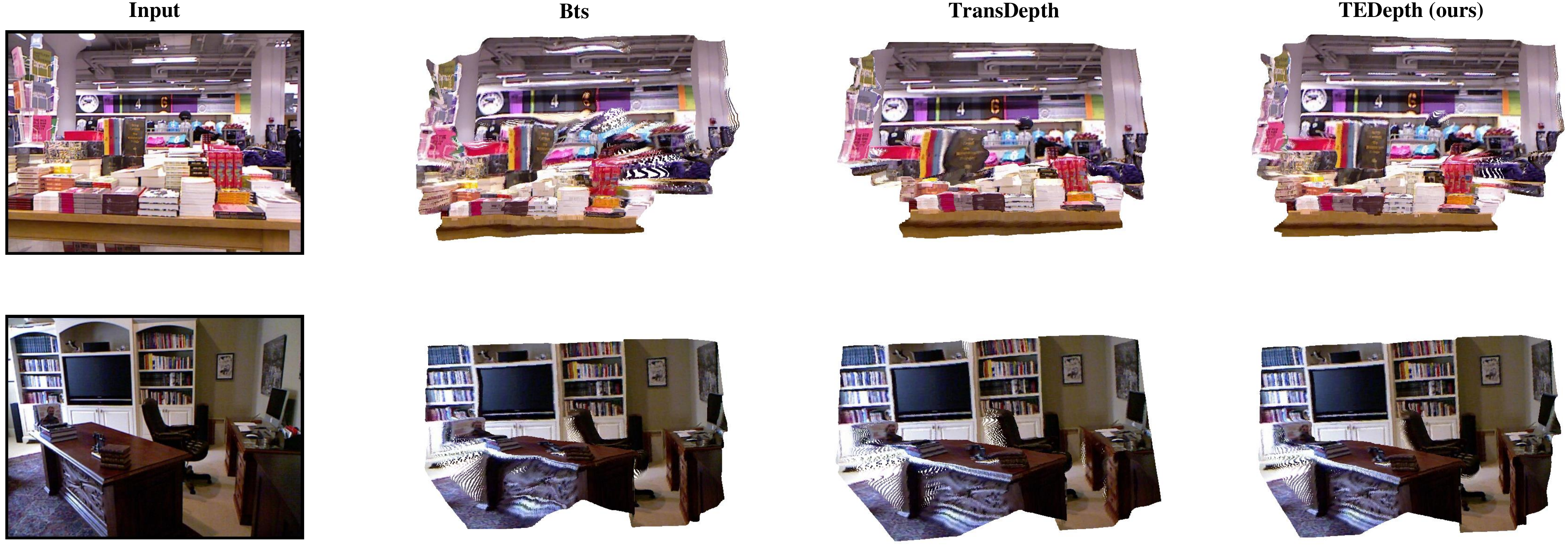}
	\caption{Visualization of reconstructed 3D scenes on the NYU-Depth-v2 dataset.}
	\label{Fig10}
\end{figure*}

\begin{figure*}[!]
	\centering
	\includegraphics[width=1.0\linewidth]{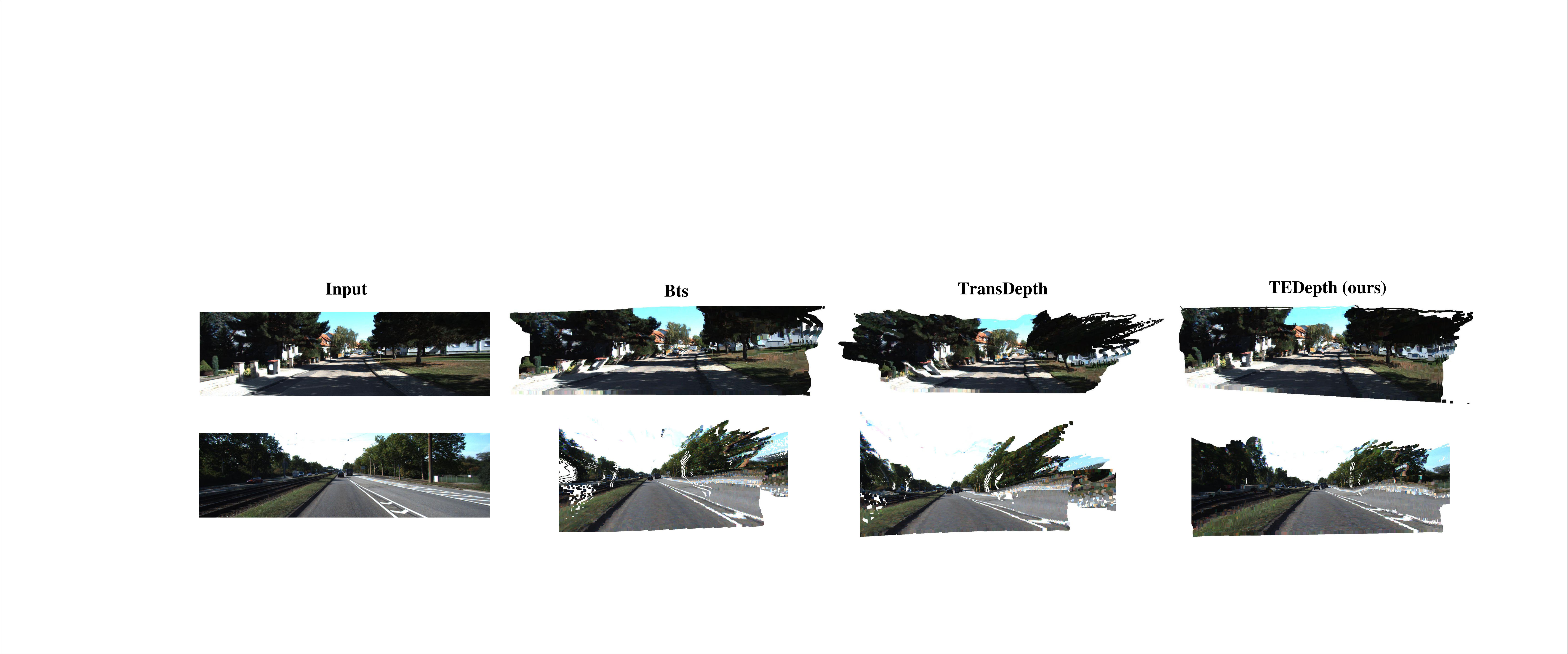}
	\caption{Visualization of reconstructed 3D scenes on the KITTI dataset.}
	\label{Fig11}
\end{figure*}

\subsection{Comparison to previous state-of-the-art competitors}
\subsubsection{Depth estimation}Table~\ref{table1} summarizes the comparison results on the NYU-Depth-v2 dataset, and TEDepth exceeds existing monocular depth estimation methods by a large margin. To demonstrate the competitiveness of our TEDepth in the outdoor scenario, we also present the comparison results on the KITTI dataset in Table~\ref{table2}. Compared to several recent competing methods such as P3Depth~\cite{patil2022p3depth} and TransDepth~\cite{yang2021transformer}, our approach is superior. 

It is worth noting that TEDepth achieves improvements on almost all metrics in both scenarios. The metrics Abs Rel, Sq Rel, RMSE, RMSE log and ${\textbf{\rm{log}}_{\bm{{10}}}}$ show the errors between the predicted depth and ground-truth, and the metrics $\delta  < 1.25$, $\delta  < {1.25^2}$ and $\delta  < {1.25^3}$ describe the number of estimated outliers. The consistent advances in reducing errors and the number of outliers support our standpoint that integrating the unique strengths of multiple weak predictor results in a more comprehensive and accurate depth predictor. 
 
Fig.~\ref{Fig3} and Fig.~\ref{Fig4} display the qualitative depth comparisons. As presented in Fig.~\ref{Fig3}, the depth maps of TEDepth deliver finer-grained details and sharper object boundaries. In Fig.~\ref{Fig4}, the compared methods struggle with thinner structures (\textit{e.g.}, signs and posts) and difficult object boundaries such as trees that overlap with foliage, while ours accurately estimates the depth of these smaller details. In addition, we visualize the reconstructed 3D scenes in Fig.~\ref{Fig10} and Fig.~\ref{Fig11}. Thanks to the comprehensive depth predictions, the point clouds of TEDepth demonstrate few distortions and preserve prominent geometric features.

\begin{figure}[!]
	\centering
	\includegraphics[width=0.95\linewidth]{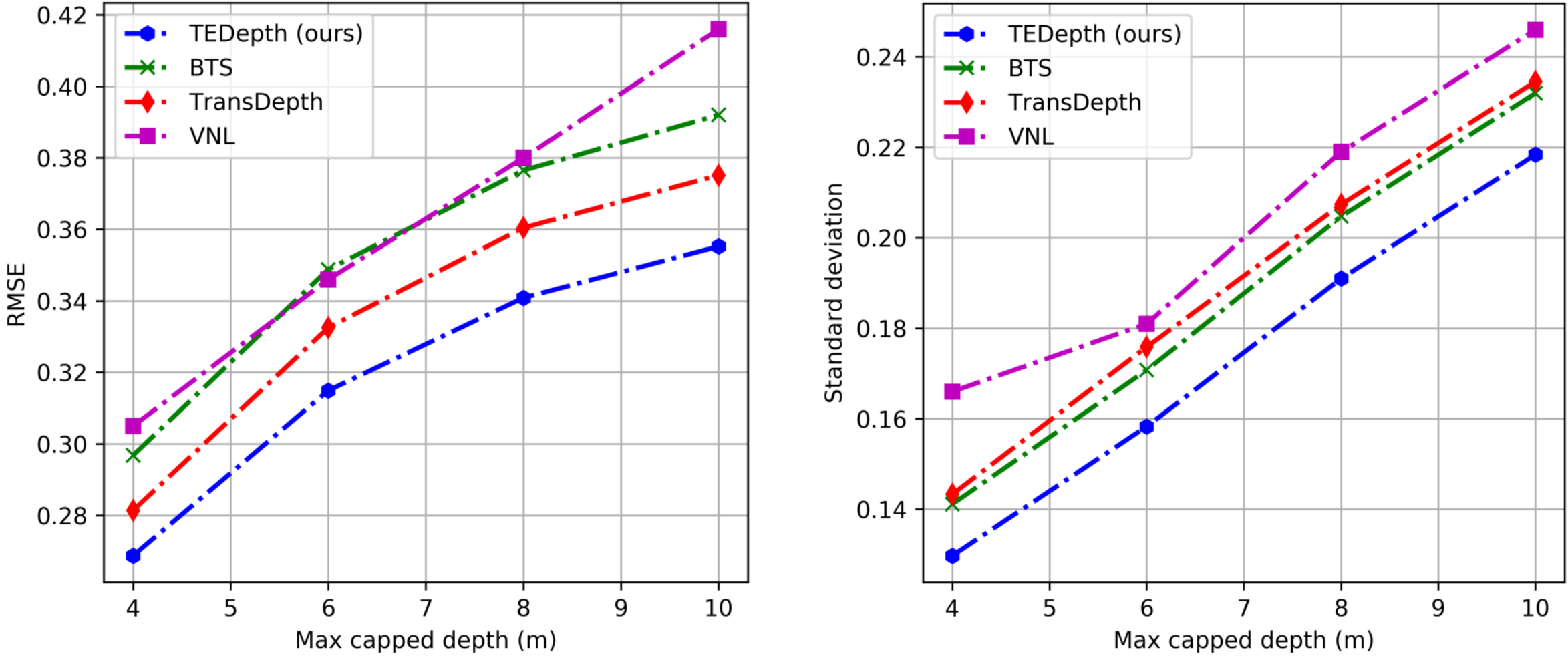}
	\caption{Illustration of the errors and corresponding standard deviations on RMSE for TransDepth, BTS, VNL and TEDepth at different capped depths on the NYU-Depth-v2 dataset. We obtain the results by clipping the predicted depth map and ground-truth within each specific depth range.}
	\label{Fig7}
\end{figure}
\begin{figure}[!]
	\centering
	\includegraphics[width=0.95\linewidth]{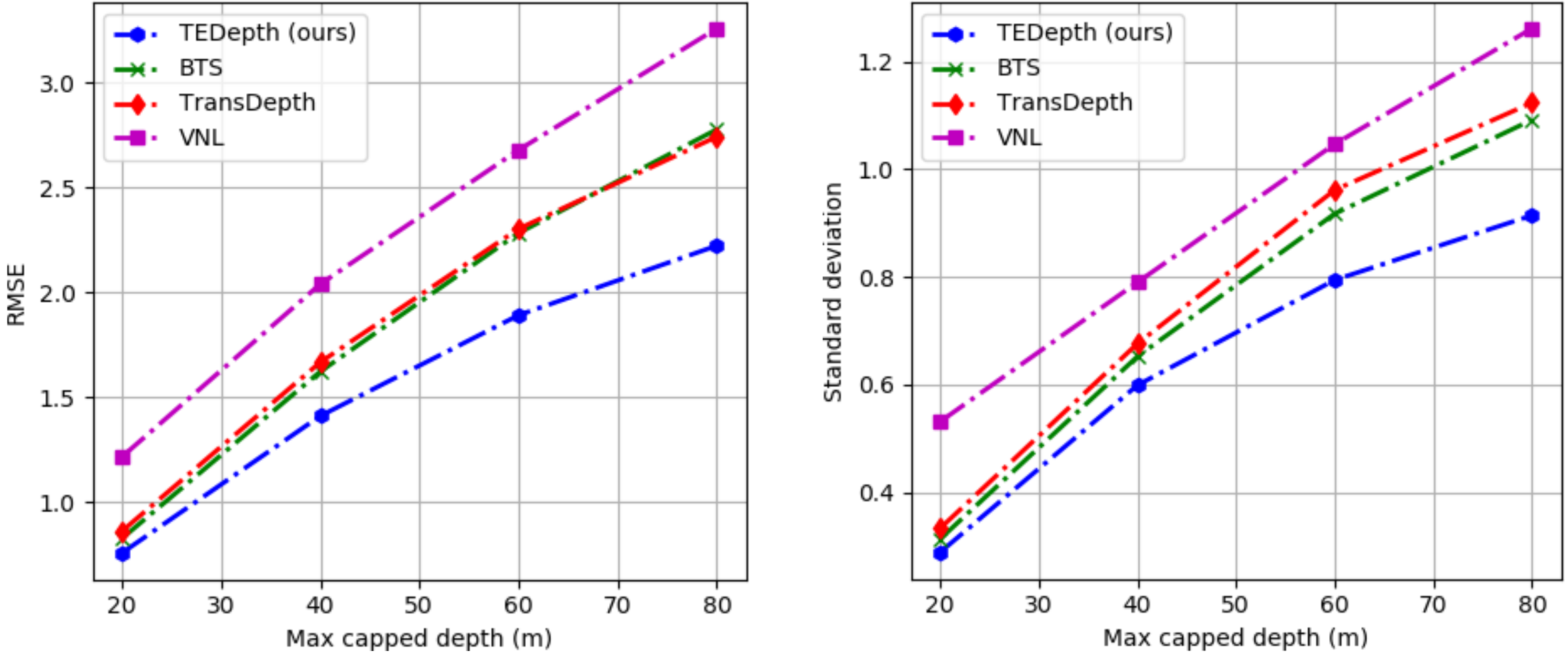}
	\caption{Illustration of the errors and corresponding standard deviations on RMSE for TransDepth, BTS, VNL and TEDepth at different capped depths on the KITTI dataset. We obtain the results by clipping the predicted depth map and ground-truth within each specific depth range.}
	\label{Fig8}
\end{figure}

Fig.~\ref{Fig7} and Fig.~\ref{Fig8} show the errors and corresponding standard deviations on the important metric RMSE at different capped depth ranges. As we can see, TransDepth is more accurate than BTS, but it appears to be less stable. By contrast, our TEDepth is able to maintain both high accuracy and stability at close, middle and distant distances.

In general, the excellent quantitative and qualitative results on NYU-Depth-v2 and KITTI datasets verify the effectiveness of the TEDepth for monocular depth estimation. 

\subsubsection{Ensemble scheme}
To further demonstrate the strengths of TEDepth, we compare it with two state-of-the-art ensemble schemes Snapshot~\cite{huang2017snapshot} and Balaji~\textit{et al.}~\cite{lakshminarayanan2016simple} in Table~\ref{table12}. Except for differences in the ensemble scheme, other configurations remain the same. We also report results of the non-ensemble method TransDepth. As can be seen, our TEDepth is able to achieve higher accuracy with fewer parameters than the compared ensemble schemes, indicating that coupling of CNN and Transformer contributes significantly to neural ensembles in monocular depth estimation. Even compared to TransDepth, TEDepth has advantages not only in terms of accuracy but also in terms of quantity of parameters,~\textit{e.g.}, in the NYU-Depth-v2 dataset. 
\begin{table*}[htb!]
	\begin{center}
		\renewcommand{\arraystretch}{1.3}
		\resizebox{1.90\columnwidth}{!}{\begin{tabular}{c || c || c c c || c c || c c}	
				\Xhline{1.2pt}
				Method & Cap & Abs Rel $\downarrow$ & Sq Rel $\downarrow$ & RMSE $\downarrow$ &  $\delta  < 1.25$ $\uparrow$ &  $\delta  < {1.25^2}$ $\uparrow$& Parameters & FLOPs \\ 
				\hline						
				\hline
				\multicolumn{9}{c}{NYU-Depth-v2}\\
				\hline
				TransDepth\FiveStar~\cite{yang2021transformer}&0-10m&0.106&-&0.365&\textbf{0.900}&0.983&247M&148G
				\\ 
				\hline 
				Snapshot~\cite{huang2017snapshot}&0-10m&0.105&0.057&0.367&0.895&\textbf{0.985}&303M&210G
				\\  
				Balaji~\textit{et al.}~\cite{lakshminarayanan2016simple}&0-10m&0.105&0.058&0.367&0.895&0.984&303M&210G
				\\ 							
				\hline
				\textbf{TEDepth (ours)}&0-10m &\textbf{0.103}&\textbf{0.054}&\textbf{0.360}&\textbf{0.900}&\textbf{0.985}&199M&170G\\
				\hline
				\hline
				\multicolumn{9}{c}{KITTI}\\
				\hline
				TransDepth\FiveStar~\cite{yang2021transformer}&0-80m&0.064&0.252&2.755&0.956&0.994&247M&207G
				\\  
				\hline
				Snapshot~\cite{huang2017snapshot}&0-80m&0.058&0.195&2.350&0.964&0.995&303M&293G
				\\  
				Balaji~\textit{et al.}~\cite{lakshminarayanan2016simple}&0-80m&0.059&0.201&2.378&0.965&0.995&303M&293G
				\\ 							
				\hline
				\textbf{TEDepth (ours)}&0-80m &\textbf{0.056}&\textbf{0.178}&\textbf{2.257}&\textbf{0.966}&\textbf{0.996}&253M&335G\\
				\Xhline{1.2pt}
		\end{tabular}}
	\end{center}
	\caption{Comparison of ensemble scheme on KITTI and NYU-Depth-v2 datasets. Note that only two base predictors are used in each ensemble scheme. Results of the non-ensemble method TransDepth are also reported, highlighted by ``\FiveStar''.  ``-'' means not applicable. The best results are indicated in bold. }
	\label{table12}
\end{table*}

\begin{table*}[htb!]
	\begin{center}
		\renewcommand{\arraystretch}{1.3}
		\resizebox{1.70\columnwidth}{!}{\begin{tabular}{c || c || c c c || c c c}	
				\Xhline{1.2pt}
				Method & Cap &  Abs Rel $\downarrow$ & RMSE $\downarrow$ & ${\textbf{\rm{log}}_{\bm{{10}}}}$ $\downarrow$ &  $\delta  < 1.25$ $\uparrow$ &  $\delta  < {1.25^2}$ $\uparrow$& $\delta  < {1.25^3}$ $\uparrow$ \\ 
				\hline						
				\hline
				BTS~\cite{lee2019big}&0-10m&0.141&0.500&0.059&0.833&0.965&0.991
				\\  
				TransDepth~\cite{yang2021transformer}&0-10m&0.149&0.472&0.059&0.875&\textbf{0.981}&0.995
				\\ 							
				\hline
				\textbf{TEDepth (ours)}&0-10m &\textbf{0.135}&\textbf{0.446}&\textbf{0.054}&\textbf{0.879}&\textbf{0.981}&\textbf{0.996}\\
				\Xhline{1.2pt}
		\end{tabular}}
	\end{center}
	\caption{Generalization on the SUN RGB-D dataset. All methods are trained on the NYU-Depth-v2 dataset. The best results are indicated in bold.}
	\label{table8}
\end{table*}

\begin{table}[htb!]
	\begin{center}
		\renewcommand{\arraystretch}{1.3}
		\resizebox{0.97\columnwidth}{!}{\begin{tabular}{c|| c c c|| c c c}	
				\Xhline{1.2pt}
				Encoder & & CNN & & &Transformer & \\
				\hline
				RMSE $\downarrow$ & top1  & top2  & top3 & top1  &  top2 & top3\\ 
				\hline						
				\hline
				\textbf{TEDepth-RBF (2)}&$\checkmark$& $-$& $-$&$\checkmark$ &$-$&$-$
				\\ 				
				\textbf{TEDepth-RBF (3)}&$\checkmark$&$-$/$\checkmark$ & $-$&$\checkmark$ & $\checkmark$/$-$ &$-$
				\\ 
				\textbf{TEDepth-RBF (4)}&$\checkmark$&$\checkmark$ & $-$&$\checkmark$ &$\checkmark$&$-$
				\\ 
				\textbf{TEDepth-RBF (5)}&$\checkmark$&$\checkmark$ &$-$/$\checkmark$ &$\checkmark$ &$\checkmark$&$\checkmark$/$-$
				\\
				\textbf{TEDepth-RBF (6)}&$\checkmark$&$\checkmark$ &$\checkmark$ &$\checkmark$ &$\checkmark$ &$\checkmark$
				\\  				
				\Xhline{1.2pt}
		\end{tabular}}
	\end{center}
	\caption{Illustration of the alternative fusing strategy. The number in parentheses stands for the number of base predictors, with the CNN-based and Transformer-based predictors fused alternately based on accuracy. ``$\checkmark$'' and ``$-$'' denotes the base predictors selected and not selected in the fusion, respectively.}
	\label{table9}
\end{table}

\begin{table}[htb!]
	\begin{center}
		\renewcommand{\arraystretch}{1.3}
		\resizebox{1.0\columnwidth}{!}{\begin{tabular}{c|| c c c || c c }	
				\Xhline{1.2pt}
				Method & Abs Rel $\downarrow$ & RMSE $\downarrow$ & ${\textbf{\rm{log}}_{\bm{{10}}}}$ $\downarrow$ & $\delta  < 1.25$ $\uparrow$ &  $\delta  < {1.25^2}$ $\uparrow$\\ 
				\hline						
				\hline
				BP-ResNet101&0.115&0.401&0.049&0.872&0.979
				\\
				BP-DenseNet161&0.111&0.392&0.047&0.879&0.981
				\\ 
				BP-R50+ViT-B/16&0.108&0.375&0.046&0.890&0.983
				\\  
				BP-Conformer&0.109&0.382&0.046&0.889&0.981
				\\
				BP-Volo&0.108&0.376&0.046&0.896&0.983
				\\
				\hline  
				\textbf{TEDepth-UWF (5)}&0.102&0.357&0.044&0.902&\textbf{0.987}
				\\ 
				\textbf{TEDepth-CGF (5)}&0.102&0.356&\textbf{0.043}&0.903&\textbf{0.987}
				\\ 
				\textbf{TEDepth-CBF (5)}&\textbf{0.100}&\textbf{0.349}&\textbf{0.043}&\textbf{0.907}&\textbf{0.987}
				\\ 
				\textbf{TEDepth-RBF (5)}&0.101&0.355&\textbf{0.043}&0.903&0.986
				\\ 		
				\Xhline{1.2pt}
		\end{tabular}}
	\end{center}
	\caption{Comparison of different mixer types on the NYU-Depth-v2 dataset. UWF: uniformly weighted fusion; CGF: confidence-guided fusion; CBF: concatenation-based fusion; RBF: ranking-based fusion. }
	\label{table5}
\end{table}

\begin{table}[htb!]
	\begin{center}
		\renewcommand{\arraystretch}{1.3}
		\resizebox{1.0\columnwidth}{!}{\begin{tabular}{c|| c c c || c c}	
				\Xhline{1.2pt}
				Method & Abs Rel $\downarrow$ & Sq Rel $\downarrow$ & RMSE $\downarrow$ & $\delta  < 1.25$ $\uparrow$ &  $\delta  < {1.25^2}$ $\uparrow$\\ 
				\hline						
				\hline
				BP-ResNext101&0.058&0.197&2.336&0.963&0.994
				\\
				BP-Conformer&0.063&0.206&2.366&0.961&0.995
				\\ 
				BP-Volo&0.069&0.219&2.378&0.959&0.995
				\\
				\hline  
				\textbf{TEDepth-UWF (3)}&0.057&0.184&2.291&0.967&\textbf{0.996}
				\\ 
				\textbf{TEDepth-CGF (3)}&\textbf{0.056}&0.182&2.293&0.967&\textbf{0.996}
				\\ 
				\textbf{TEDepth-CBF (3)}&0.057&0.179&2.236&\textbf{0.968}&\textbf{0.996}
				\\ 
				\textbf{TEDepth-RBF (3)}&\textbf{0.056}&\textbf{0.174}&\textbf{2.223}&\textbf{0.968}&\textbf{0.996}
				\\ 				
				\Xhline{1.2pt}
		\end{tabular}}
	\end{center}
	\caption{Comparison of different design choices for mixers on the KITTI dataset. }
	\label{table6}
\end{table}

\begin{table}[htb!]
	\begin{center}
		\renewcommand{\arraystretch}{1.3}
		\resizebox{1.0\columnwidth}{!}{\begin{tabular}{c|| c c c || c c}	
				\Xhline{1.2pt}
				\hline	
				Method  & Abs Rel $\downarrow$ & Sq Rel $\downarrow$ & RMSE $\downarrow$ & $\delta  < 1.25$ $\uparrow$ &  $\delta  < {1.25^2}$ $\uparrow$\\ 
				\hline						
				\hline
				\multicolumn{6}{c}{NYU-Depth-v2}\\
				\hline
				\textbf{BP-ResNet101$\&$BP-DenseNet161 }&0.108&0.060&0.381&0.887&0.983
				\\ 								
				\textbf{(BP-R50+ViT-B/16)$\&$BP-Volo }&0.104&0.056&0.363&0.897&\textbf{0.985}
				\\ 
				\textbf{(BP-R50+ViT-B/16)$\times$2 }&0.105&0.058&0.367&0.895&0.984
				\\ 
				\textbf{TEDepth-RBF (2)}&\textbf{0.103}&\textbf{0.054}&\textbf{0.360}&\textbf{0.900}&\textbf{0.985}
				\\ 
				\hline
				\hline
				\multicolumn{6}{c}{KITTI}\\
				\hline	
				\textbf{BP-ResNext101$\&$BP-DenseNet161 }&\textbf{0.055}&0.184&2.313&0.965&0.995
				\\ 								
				\textbf{BP-Conformer$\&$BP-Volo }&0.058&0.191&2.317&0.964&0.995
				\\
				\textbf{(BP-ResNext101)$\times$2 }&0.058&0.198&2.363&0.963&0.994
				\\  
				\textbf{TEDepth-RBF (2)}&0.056&\textbf{0.178}&\textbf{2.257}&\textbf{0.966}&\textbf{0.996}
				\\ 				
				\Xhline{1.2pt}
		\end{tabular}}
	\end{center}
	\caption{Effect of diversity among base predictors. All variants adopt the same mixer RBF.}
	\label{table7}
\end{table}
\subsection{Generalization on the SUN RGB-D dataset}
We also study the generalizability of TEDepth, as evidence that TEDepth does learn transferable features rather than simply memorize training data. To verify this, we evaluate the models trained by NYU-Depth-v2 on the SUN RGB-D dataset without any fine-tuning. As shown in Table~\ref{table8}, TransDepth achieves a significant reduction in root mean square error and the number of estimated outliers compared to BTS, but a larger error on the Abs Rel, which is susceptible to the close-range errors. In contrast, TEDepth generalizes well on all metrics across different cameras. 
\begin{table*}[htb!]
	\begin{center}
		\renewcommand{\arraystretch}{1.3}
		\resizebox{1.85\columnwidth}{!}{\begin{tabular}{c|| c c c || c c || c c }	
				\Xhline{1.2pt}
				Method & Abs Rel $\downarrow$ & RMSE $\downarrow$ & ${\textbf{\rm{log}}_{\bm{{10}}}}$ $\downarrow$ & $\delta  < 1.25$ $\uparrow$ &  $\delta  < {1.25^2}$ $\uparrow$ & Parameters & FLOPs\\ 
				\hline						
				\hline
				BP-ResNet101&0.115&0.401&0.049&0.872&0.979&69M&66G
				\\
				BP-DenseNet161&0.111&0.392&0.047&0.879&0.981&47M&61G
				\\
				BP-ResNext101&0.114&0.402&0.049&0.873&0.977&113M&93G
				\\ 
				BP-R50+ViT-B/16&0.108&0.375&0.046&0.890&0.983&152M&101G
				\\  
				BP-Conformer&0.109&0.382&0.046&0.889&0.981&141M&139G
				\\
				BP-Volo&0.108&0.376&0.046&0.896&0.983&128M&117G
				\\
				\hline  
				\textbf{TEDepth-RBF (2)}&0.103&0.360&0.044&0.900&0.985&199M&170G
				\\ 
				\textbf{TEDepth-RBF (3)}&0.102&0.358&0.044&0.901&\textbf{0.986}&326M&291G
				\\ 
				\textbf{TEDepth-RBF (4)}&0.102&\textbf{0.355}&0.044&\textbf{0.903}&\textbf{0.986}&395M&362G
				\\ 
				\textbf{TEDepth-RBF (5)}&\textbf{0.101}&\textbf{0.355}&\textbf{0.043}&\textbf{0.903}&\textbf{0.986}&535M&505G
				\\
				\textbf{TEDepth-RBF (6)}&\textbf{0.101}&\textbf{0.355}&\textbf{0.043}&0.902&\textbf{0.986}&648M&602G
				\\ 
				\Xhline{1.2pt}
		\end{tabular}}
	\end{center}
	\caption{Effect of the number of base predictors on the NYU-Depth-v2 dataset. BP: base predictor; RBF: ranking-based fusion. The detailed explanations from TEDepth-RBF (2) to TEDepth-RBF (6) are in Table~\ref{table9}.} 
	\label{table3}
\end{table*}

\begin{table*}[htb!]
	\begin{center}
		\renewcommand{\arraystretch}{1.3}
		\resizebox{1.85\columnwidth}{!}{\begin{tabular}{c|| c c c || c c || c c }	
				\Xhline{1.2pt}
				Method & Abs Rel $\downarrow$ & Sq Rel $\downarrow$ & RMSE $\downarrow$ & $\delta  < 1.25$ $\uparrow$ &  $\delta  < {1.25^2}$ $\uparrow$ & Parameters & FLOPs\\ 
				\hline						
				\hline
				BP-ResNet101&0.061&0.213&2.420&0.961&0.994&69M&92G
				\\
				BP-DenseNet161&0.058&0.201&2.420&0.962&0.994&47M&85G
				\\
				BP-ResNext101&0.058&0.197&2.336&0.963&0.994&113M&129G
				\\ 
				BP-R50+ViT-B/16&0.061&0.213&2.409&0.962&0.995&152M&140G
				\\  
				BP-Conformer&0.063&0.206&2.366&0.961&0.995&141M&194G
				\\
				BP-Volo&0.069&0.219&2.378&0.959&0.995&128M&163G
				\\
				\hline  
				\textbf{TEDepth-RBF (2)}&\textbf{0.056}&0.178&2.257&0.966&0.996&254M&335G
				\\ 
				\textbf{TEDepth-RBF (3)}&\textbf{0.056}&\textbf{0.174}&\textbf{2.223}&\textbf{0.968}&\textbf{0.996}&382M&503G
				\\ 
				\textbf{TEDepth-RBF (4)}&0.057&0.185&2.271&0.967&\textbf{0.996}&429M&594G
				\\ 
				\textbf{TEDepth-RBF (5)}&\textbf{0.056}&0.182&2.265&\textbf{0.968}&\textbf{0.996}&676M&740G
				\\ 
				\textbf{TEDepth-RBF (6)}&0.057&0.183&2.254&\textbf{0.968}&\textbf{0.996}&745M&838G
				\\ 
				\Xhline{1.2pt}
		\end{tabular}}
	\end{center}
	\caption{Effect of the number of base predictors on the KITTI dataset. }
	\label{table4}
\end{table*}

\begin{table}[htb!]
	\begin{center}
		\renewcommand{\arraystretch}{1.3}
		\resizebox{1.0\columnwidth}{!}{\begin{tabular}{c|| c c ||c c}	
				\Xhline{1.2pt}
				Dataset&\multicolumn{2}{c||}{NYU-Depth-v2}&\multicolumn{2}{c}{KITTI}\\
				\hline
				Variant &FL-RBF (5) &PL-RBF (5)  &FL-RBF (3) &PL-RBF (3)   \\ 
				\hline						
				\hline
				Abs Rel $\downarrow$  &0.116&\textbf{0.101}&0.068&\textbf{0.056}
				\\ 						
				Sq Rel $\downarrow$&0.067&\textbf{0.052}&0.237&\textbf{0.174}
				\\ 
				RMSE $\downarrow$&0.403&\textbf{0.355}&2.534&\textbf{2.223}
				\\ 
				$\log_{10}$ $\downarrow$  &0.050&\textbf{0.043}&0.030&\textbf{0.024}
				\\ 				
				$\delta  < 1.25$ $\uparrow$&0.868&\textbf{0.903}&0.956&\textbf{0.968}
				\\ 		
				$\delta  < {1.25^2}$$\uparrow$&0.979&\textbf{0.986}&0.993&\textbf{0.996}
				\\ 
				$\delta  < {1.25^3}$$\uparrow$&0.996&\textbf{0.998}&0.993&\textbf{0.999}
				\\	
				\Xhline{1.2pt}
		\end{tabular}}
	\end{center}
	\caption{Comparison of fusing at penultimate and final layers. FL: final layer, PL: penultimate layer.}
	\label{table11}
\end{table}
\subsection{Ablation studies}
To better understand how the different elements in TEDepth impair the overall performance, we perform a series of ablation studies, involving the number of base predictors, mixer types, diversity among base predictors and fusion locations. 

\subsubsection{Number of base predictors} 
Tables~\ref{table3} and~\ref{table4} demonstrate the impact of number of base predictors on accuracy, parameters and FLOPs. We adopt the ranking-based fusion as mixer. It is interesting that after a certain number of base predictors fused, ensemble accuracy stops improving or even declines on some metrics. The phenomenon is consistent with the findings in~\cite{zhou2002ensembling} that many could be better than all. In addition, we notice that when the two base predictors are integrated, the accuracy improvement is the greatest, while the parameters and FLOPs of TEDepth are smallest. The increase in accuracy becomes smaller and smaller as the base predictors are fused further. Therefore, for new datasets, we advocate integrating two or three base predictors to get a good trade-off between performance and efficiency. 

\subsubsection{Mixer types} In Tables~\ref{table5} and~\ref{table6}, we investigate the influence of different mixer types. It is clear that, regardless of the deployment type, TEDepth can significantly improve accuracy over base predictors. As a result, in the case of ensuring diversity among base predictors, the ensemble operation determines the approximate range of accuracy. Besides, the uniformly weighted fusion mixer performs the worst out of the four types of mixers, and the concatenation-based and ranking-based fusions perform the best on NYU-Depth-v2 and KITTI, respectively.

\subsubsection{Diversity among base predictors} Table~\ref{table7} summarizes the effect of diversity among base predictors. We compare TEDepth-RBF (2) with the best two Transformer-based fusion model, the best two CNN-based fusion model and the best two identical base predictors fusion model. Thanks to the complementarity of CNN and visual transformer, our TEDepth-RBF (2) performs better on two datasets.

\subsubsection{Fusion locations} As shown in Table~\ref{table11}, we present the comparison of fusing at penultimate and final layers. Given that the multi-channel feature maps contain more exploitable information than the single-channel depth maps, the results verify the deduction. 

\begin{figure*}[!]
	\centering
	\includegraphics[width=1.0\linewidth]{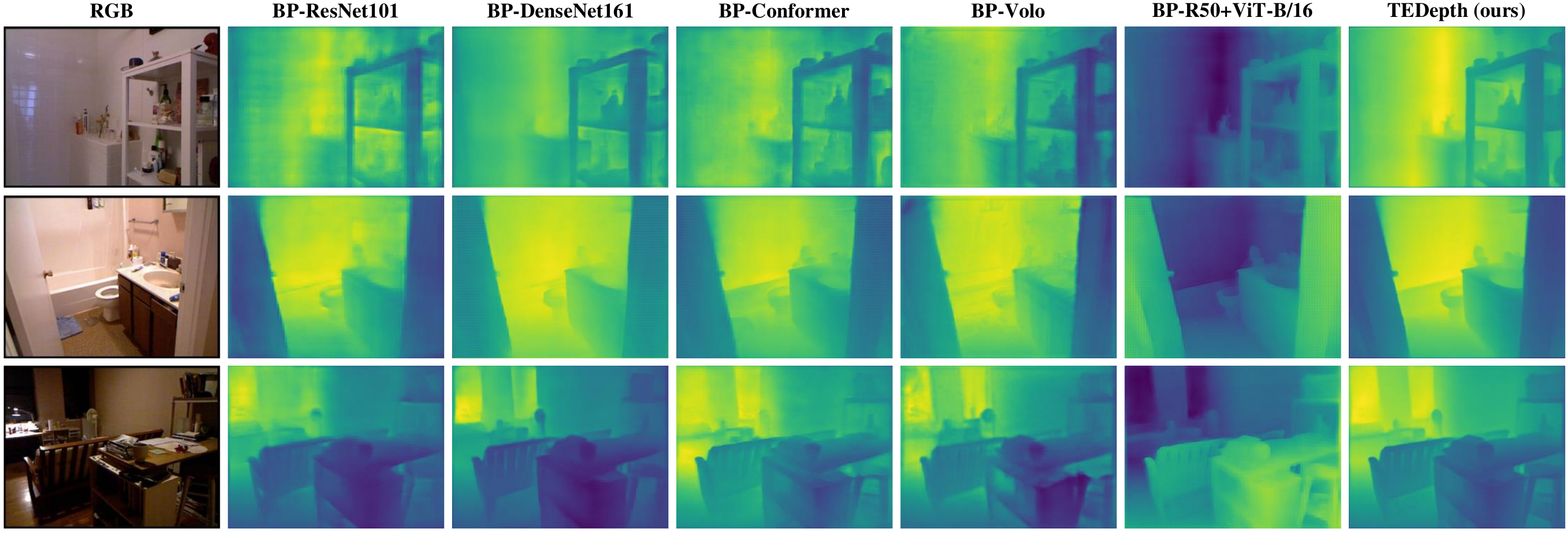}
	\caption{Visualization of the penultimate layer feature maps on the NYU-Depth-v2 dataset, achieved by choosing one principle channel via PCA decomposition and then displaying the feature map as a heat map. }
	\label{Fig5}
\end{figure*}

\begin{figure*}[!]
	\centering
	\includegraphics[width=1.0\linewidth]{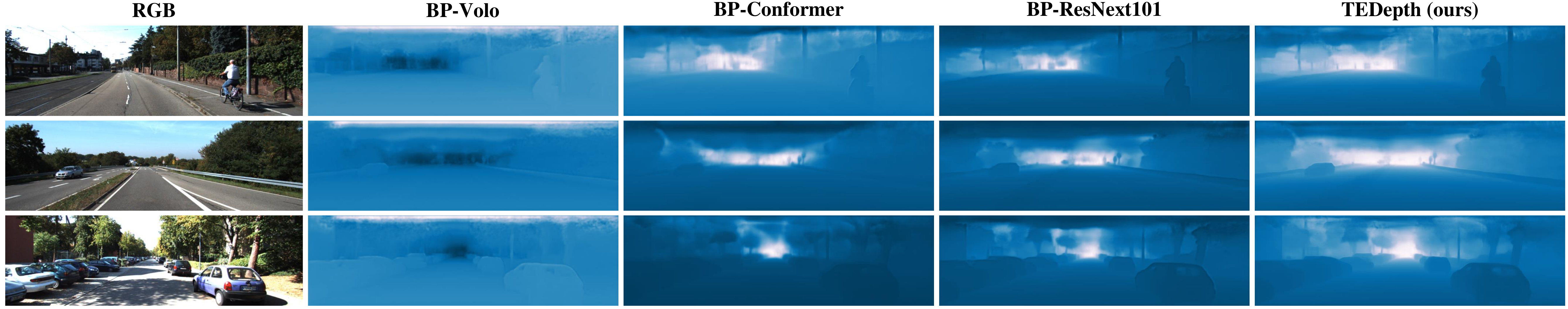}
	\caption{Visualization of the penultimate layer feature maps from BP-Volo, BP-Conformer, BP-ResNext101 and TEDepth on the KITTI dataset.}
	\label{Fig6}
\end{figure*}

In Fig.~\ref{Fig5} and Fig.~\ref{Fig6}, we show the penultimate layer feature maps from base predictors to assess their contribution towards the asymmetric depth errors and the feature maps of TEDepth to highlight how fusing multiple predictors handles asymmetric depth errors. We find that the chosen base predictors indeed provide asymmetric feature maps, and the TEDepth is able to inherit the strengths of each base predictor to alleviate their catastrophic errors.

\begin{figure}[!]
	\centering
	\includegraphics[width=1.0\linewidth]{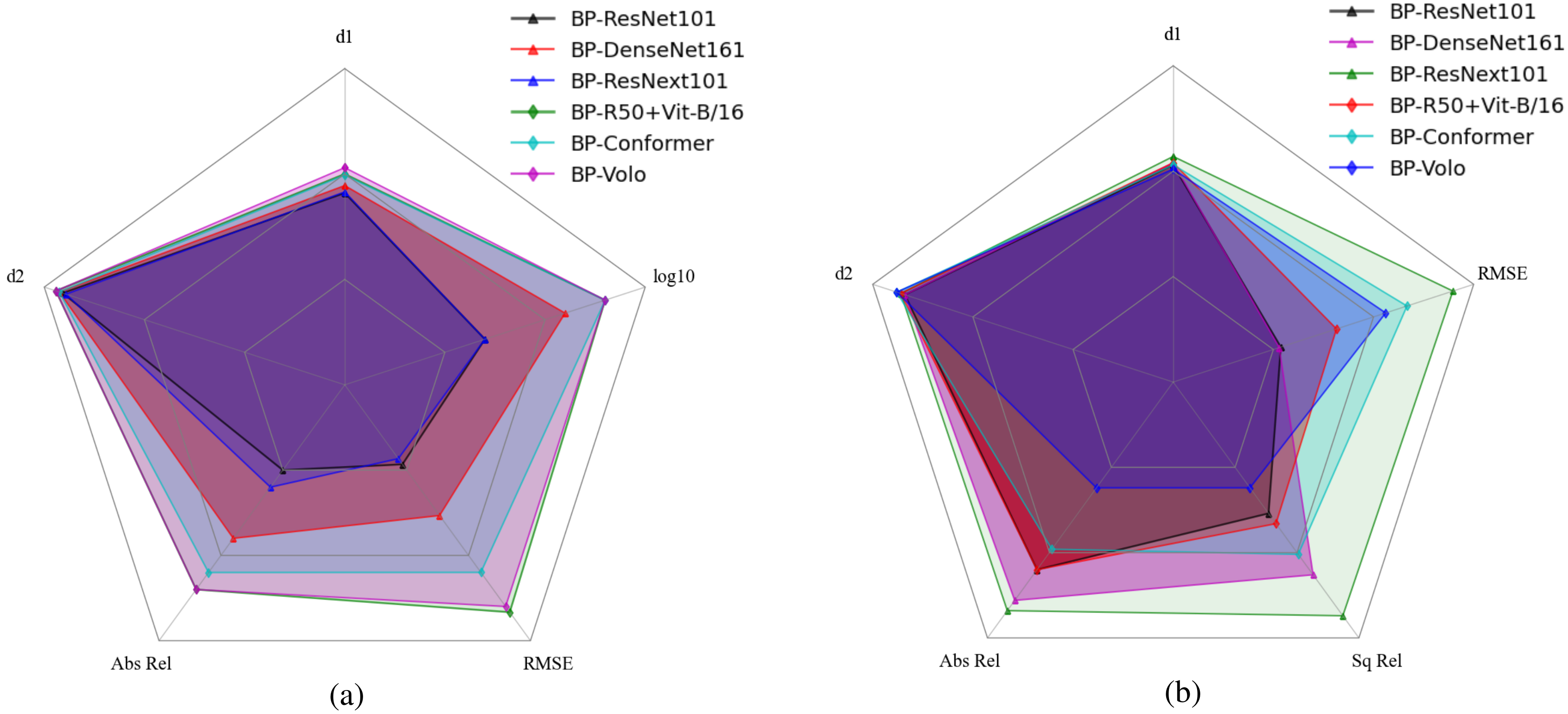}
	\caption{Comparison of performance with respect to the base predictors. It should be noted that the larger the shadow area, the smaller the model errors. (a) Comparison on the NYU-Depth-v2 dataset. (b) Comparison on the KITTI dataset.}
	\label{Fig9}
\end{figure}

In Fig.~\ref{Fig9}, we present the performance comparison of base predictors. An interesting phenomenon is that Transformer-based predictors perform better on NYU-Depth-v2, whereas CNN-based predictors perform better on KITTI. One reason for this could be the ground-truth type. NYU-Depth-v2 has dense ground-truth depth maps. On the other hand, the ground-truth depth maps of KITTI are relatively sparse. By capturing long-range dependencies, visual transformer can improve accuracy at pixels without ground-truth supervision. It may, however, degrade the depth accuracy at pixels supervised by ground-truth. Because of the dense ground-truth depth maps in the NYU-Depth-v2 dataset, it is easier to obtain more accurate depth maps through long-range dependencies. In the case of the KITTI dataset, the opposite is true.


\section{Conclusion}
In this work, we open up a new avenue for model design by introducing ensemble learning and thoroughly investigate its utility for monocular depth estimation. Besides, we develop an effective and easy-to-implement framework, TEDepth, based on two levels to deliver asymmetric bottom-level predictions and adaptively merge them at the top level, enabling a comprehensive and accurate depth estimation. Extensive experiments are carried out on three challenging datasets, including NYU-Depth-v2, KITTI and SUN RGB-D datasets. The experimental results emphasize the effectiveness and strong generalizability of TEDepth. We hope our study can encourage more works applying the neural ensembles into monocular depth estimation and enlighten the framework design of other closely related tasks,~\textit{e.g.}, surface normal prediction.

\section{CRediT authorship contribution statement}
Shuwei Shao: Conceptualization, Methodology, Software, Validation, Writing - original draft, Writing - review $\&$ editing. Ran Li: Methodology, Software, Validation, Investigation, Visualization, Writing - review $\&$ editing. Zhongcai Pei: Software, Validation, Investigation, Writing - review $\&$ editing. Zhong Liu: Software, Validation, Visualization, Writing - review $\&$ editing. Weihai Chen: Resources, Funding acquisition, Writing - review $\&$ editing, Supervision. Wentao Zhu: Software, Validation, Writing - review $\&$ editing. Xingming Wu: Investigation, Writing - review $\&$ editing. Baochang Zhang: Conceptualization, Methodology, Resources, Writing - review $\&$ editing, Supervision.
\normalem

{\small
	\bibliographystyle{unsrt}
	\bibliography{reference}

\begin{thebibliography}{10}

\bibitem{lee2019big}
Jin~Han Lee, Myung-Kyu Han, Dong~Wook Ko, and Il~Hong Suh.
\newblock From big to small: Multi-scale local planar guidance for monocular
  depth estimation.
\newblock {\em arXiv preprint arXiv:1907.10326}, 2019.

\bibitem{yang2021transformer}
Guanglei Yang, Hao Tang, Mingli Ding, Nicu Sebe, and Elisa Ricci.
\newblock Transformer-based attention networks for continuous pixel-wise
  prediction.
\newblock In {\em Proceedings of the IEEE International Conference on Computer
  Vision}, pages 16269--16279, October 2021.

\bibitem{hazirbas2016fusenet}
Caner Hazirbas, Lingni Ma, Csaba Domokos, and Daniel Cremers.
\newblock Fusenet: Incorporating depth into semantic segmentation via
  fusion-based cnn architecture.
\newblock In {\em Asian Conference on Computer Vision}, pages 213--228.
  Springer, 2016.

\bibitem{lee2011depth}
Wonwoo Lee, Nohyoung Park, and Woontack Woo.
\newblock Depth-assisted real-time 3d object detection for augmented reality.
\newblock In {\em ICAT}, volume 11 (2), pages 126--132, 2011.

\bibitem{shao2022self}
Shuwei Shao, Zhongcai Pei, Weihai Chen, Wentao Zhu, Xingming Wu, Dianmin Sun,
  and Baochang Zhang.
\newblock Self-supervised monocular depth and ego-motion estimation in
  endoscopy: Appearance flow to the rescue.
\newblock {\em Medical image analysis}, 77:102338, 2022.

\bibitem{saxena2005learning}
Ashutosh Saxena, Sung~H Chung, Andrew~Y Ng, et~al.
\newblock Learning depth from single monocular images.
\newblock In {\em Advances in Neural Information Processing Systems},
  volume~18, pages 1--8, 2005.

\bibitem{eigen2015predicting}
David Eigen and Rob Fergus.
\newblock Predicting depth, surface normals and semantic labels with a common
  multi-scale convolutional architecture.
\newblock In {\em Proceedings of the IEEE International Conference on Computer
  Vision}, pages 2650--2658, 2015.

\bibitem{laina2016deeper}
Iro Laina, Christian Rupprecht, Vasileios Belagiannis, Federico Tombari, and
  Nassir Navab.
\newblock Deeper depth prediction with fully convolutional residual networks.
\newblock In {\em 2016 Fourth International Conference on 3D Vision}, pages
  239--248. IEEE, 2016.

\bibitem{li2017two}
Jun Li, Reinhard Klein, and Angela Yao.
\newblock A two-streamed network for estimating fine-scaled depth maps from
  single rgb images.
\newblock In {\em Proceedings of the IEEE International Conference on Computer
  Vision}, pages 3372--3380, 2017.

\bibitem{fu2018deep}
Huan Fu, Mingming Gong, Chaohui Wang, Kayhan Batmanghelich, and Dacheng Tao.
\newblock Deep ordinal regression network for monocular depth estimation.
\newblock In {\em Proceedings of the IEEE Conference on Computer Vision and
  Pattern Recognition}, pages 2002--2011, 2018.

\bibitem{song2019contextualized}
Wenfeng Song, Shuai Li, Ji~Liu, Aimin Hao, Qinping Zhao, and Hong Qin.
\newblock Contextualized cnn for scene-aware depth estimation from single rgb
  image.
\newblock {\em IEEE Transactions on Multimedia}, 22(5):1220--1233, 2019.

\bibitem{yang2019bayesian}
Xin Yang, Yang Gao, Hongcheng Luo, Chunyuan Liao, and Kwang-Ting Cheng.
\newblock Bayesian denet: Monocular depth prediction and frame-wise fusion with
  synchronized uncertainty.
\newblock {\em IEEE Transactions on Multimedia}, 21(11):2701--2713, 2019.

\bibitem{ling2021unsupervised}
Chuanwu Ling, Xiaogang Zhang, and Hua Chen.
\newblock Unsupervised monocular depth estimation using attention and
  multi-warp reconstruction.
\newblock {\em IEEE Transactions on Multimedia}, 2021.

\bibitem{silberman2012indoor}
Nathan Silberman, Derek Hoiem, Pushmeet Kohli, and Rob Fergus.
\newblock Indoor segmentation and support inference from rgbd images.
\newblock In {\em European Conference on Computer Vision}, pages 746--760.
  Springer, 2012.

\bibitem{dong2020survey}
Xibin Dong, Zhiwen Yu, Wenming Cao, Yifan Shi, and Qianli Ma.
\newblock A survey on ensemble learning.
\newblock {\em Frontiers of Computer Science}, 14(2):241--258, 2020.

\bibitem{huang2017snapshot}
Gao Huang, Yixuan Li, Geoff Pleiss, Zhuang Liu, John~E Hopcroft, and Kilian~Q
  Weinberger.
\newblock Snapshot ensembles: Train 1, get m for free.
\newblock {\em International Conference on Learning Representations}, 2017.

\bibitem{yang2021local}
Yongquan Yang, Haijun Lv, Ning Chen, Yang Wu, Jiayi Zheng, and Zhongxi Zheng.
\newblock Local minima found in the subparameter space can be effective for
  ensembles of deep convolutional neural networks.
\newblock {\em Pattern Recognition}, 109:107582, 2021.

\bibitem{solovyev2019weighted}
Roman Solovyev, Weimin Wang, and Tatiana Gabruseva.
\newblock Weighted boxes fusion: Ensembling boxes from different object
  detection models.
\newblock {\em Image and Vision Computing}, 107:104117, 2021.

\bibitem{chen2017deeplab}
Liang-Chieh Chen, George Papandreou, Iasonas Kokkinos, Kevin Murphy, and Alan~L
  Yuille.
\newblock Deeplab: Semantic image segmentation with deep convolutional nets,
  atrous convolution, and fully connected crfs.
\newblock {\em IEEE Transactions on Pattern Analysis and Machine Intelligence},
  40(4):834--848, 2017.

\bibitem{ranftl2021vision}
Ren{\'e} Ranftl, Alexey Bochkovskiy, and Vladlen Koltun.
\newblock Vision transformers for dense prediction.
\newblock In {\em Proceedings of the IEEE International Conference on Computer
  Vision}, pages 12179--12188, 2021.

\bibitem{yuan2021volo}
Li~Yuan, Qibin Hou, Zihang Jiang, Jiashi Feng, and Shuicheng Yan.
\newblock Volo: Vision outlooker for visual recognition.
\newblock {\em arXiv preprint arXiv:2106.13112}, 2021.

\bibitem{Peng_2021_ICCV}
Zhiliang Peng, Wei Huang, Shanzhi Gu, Lingxi Xie, Yaowei Wang, Jianbin Jiao,
  and Qixiang Ye.
\newblock Conformer: Local features coupling global representations for visual
  recognition.
\newblock In {\em Proceedings of the IEEE International Conference on Computer
  Vision}, pages 367--376, October 2021.

\bibitem{dosovitskiy2020image}
Alexey Dosovitskiy, Lucas Beyer, Alexander Kolesnikov, Dirk Weissenborn,
  Xiaohua Zhai, Thomas Unterthiner, Mostafa Dehghani, Matthias Minderer, Georg
  Heigold, Sylvain Gelly, et~al.
\newblock An image is worth 16x16 words: Transformers for image recognition at
  scale.
\newblock {\em International Conference on Learning Representations}, 2021.

\bibitem{He_2016_CVPR}
Kaiming He, Xiangyu Zhang, Shaoqing Ren, and Jian Sun.
\newblock Deep residual learning for image recognition.
\newblock In {\em Proceedings of the IEEE Conference on Computer Vision and
  Pattern Recognition}, June 2016.

\bibitem{huang2017densely}
Gao Huang, Zhuang Liu, Laurens Van Der~Maaten, and Kilian~Q Weinberger.
\newblock Densely connected convolutional networks.
\newblock In {\em Proceedings of the IEEE Conference on Computer Vision and
  Pattern Recognition}, pages 4700--4708, 2017.

\bibitem{xie2017aggregated}
Saining Xie, Ross Girshick, Piotr Doll{\'a}r, Zhuowen Tu, and Kaiming He.
\newblock Aggregated residual transformations for deep neural networks.
\newblock In {\em Proceedings of the IEEE Conference on Computer Vision and
  Pattern Recognition}, pages 1492--1500, 2017.

\bibitem{song2015sun}
Shuran Song, Samuel~P Lichtenberg, and Jianxiong Xiao.
\newblock Sun rgb-d: A rgb-d scene understanding benchmark suite.
\newblock In {\em Proceedings of the IEEE Conference on Computer Vision and
  Pattern Recognition}, pages 567--576, 2015.

\bibitem{geiger2013vision}
Andreas Geiger, Philip Lenz, Christoph Stiller, and Raquel Urtasun.
\newblock Vision meets robotics: The kitti dataset.
\newblock {\em The International Journal of Robotics Research},
  32(11):1231--1237, 2013.

\bibitem{eigen2014depth}
David Eigen, Christian Puhrsch, and Rob Fergus.
\newblock Depth map prediction from a single image using a multi-scale deep
  network.
\newblock In {\em Advances in Neural Information Processing Systems}, pages
  2366--2374, 2014.

\bibitem{cao2017estimating}
Yuanzhouhan Cao, Zifeng Wu, and Chunhua Shen.
\newblock Estimating depth from monocular images as classification using deep
  fully convolutional residual networks.
\newblock {\em IEEE Transactions on Circuits and Systems for Video Technology},
  28(11):3174--3182, 2017.

\bibitem{qi2018geonet}
Xiaojuan Qi, Renjie Liao, Zhengzhe Liu, Raquel Urtasun, and Jiaya Jia.
\newblock Geonet: Geometric neural network for joint depth and surface normal
  estimation.
\newblock In {\em Proceedings of the IEEE Conference on Computer Vision and
  Pattern Recognition}, pages 283--291, 2018.

\bibitem{yin2019enforcing}
Wei Yin, Yifan Liu, Chunhua Shen, and Youliang Yan.
\newblock Enforcing geometric constraints of virtual normal for depth
  prediction.
\newblock In {\em Proceedings of the IEEE International Conference on Computer
  Vision}, pages 5684--5693, 2019.

\bibitem{patil2022p3depth}
Vaishakh Patil, Christos Sakaridis, Alexander Liniger, and Luc Van~Gool.
\newblock P3depth: Monocular depth estimation with a piecewise planarity prior.
\newblock In {\em Proceedings of the IEEE Conference on Computer Vision and
  Pattern Recognition}, pages 1610--1621, 2022.

\bibitem{vaswani2017attention}
Ashish Vaswani, Noam Shazeer, Niki Parmar, Jakob Uszkoreit, Llion Jones,
  Aidan~N Gomez, {\L}ukasz Kaiser, and Illia Polosukhin.
\newblock Attention is all you need.
\newblock In {\em Advances in Neural Information Processing Systems}, pages
  5998--6008, 2017.

\bibitem{Liu_2021_ICCV}
Ze~Liu, Yutong Lin, Yue Cao, Han Hu, Yixuan Wei, Zheng Zhang, Stephen Lin, and
  Baining Guo.
\newblock Swin transformer: Hierarchical vision transformer using shifted
  windows.
\newblock In {\em Proceedings of the IEEE International Conference on Computer
  Vision}, pages 10012--10022, October 2021.

\bibitem{zheng2021rethinking}
Sixiao Zheng, Jiachen Lu, Hengshuang Zhao, Xiatian Zhu, Zekun Luo, Yabiao Wang,
  Yanwei Fu, Jianfeng Feng, Tao Xiang, Philip~HS Torr, et~al.
\newblock Rethinking semantic segmentation from a sequence-to-sequence
  perspective with transformers.
\newblock In {\em Proceedings of the IEEE Conference on Computer Vision and
  Pattern Recognition}, pages 6881--6890, 2021.

\bibitem{gao2021ts}
Wei Gao, Fang Wan, Xingjia Pan, Zhiliang Peng, Qi~Tian, Zhenjun Han, Bolei
  Zhou, and Qixiang Ye.
\newblock Ts-cam: Token semantic coupled attention map for weakly supervised
  object localization.
\newblock In {\em Proceedings of the IEEE/CVF International Conference on
  Computer Vision}, pages 2886--2895, 2021.

\bibitem{jiang2021transgan}
Yifan Jiang, Shiyu Chang, and Zhangyang Wang.
\newblock Transgan: Two transformers can make one strong gan.
\newblock {\em Advances in Neural Information Processing Systems}, 2021.

\bibitem{bauer1999empirical}
Eric Bauer and Ron Kohavi.
\newblock An empirical comparison of voting classification algorithms: Bagging,
  boosting, and variants.
\newblock {\em Machine learning}, 36(1):105--139, 1999.

\bibitem{zhou2002ensembling}
Zhi-Hua Zhou, Jianxin Wu, and Wei Tang.
\newblock Ensembling neural networks: many could be better than all.
\newblock {\em Artificial intelligence}, 137(1-2):239--263, 2002.

\bibitem{lakshminarayanan2016simple}
Balaji Lakshminarayanan, Alexander Pritzel, and Charles Blundell.
\newblock Simple and scalable predictive uncertainty estimation using deep
  ensembles.
\newblock {\em Advances in Neural Information Processing Systems}, pages
  6402--6413, 2017.

\bibitem{wenzel2020hyperparameter}
Florian Wenzel, Jasper Snoek, Dustin Tran, and Rodolphe Jenatton.
\newblock Hyperparameter ensembles for robustness and uncertainty
  quantification.
\newblock {\em arXiv preprint arXiv:2006.13570}, 2020.

\bibitem{yang2020dverge}
Huanrui Yang, Jingyang Zhang, Hongliang Dong, Nathan Inkawhich, Andrew Gardner,
  Andrew Touchet, Wesley Wilkes, Heath Berry, and Hai Li.
\newblock Dverge: Diversifying vulnerabilities for enhanced robust generation
  of ensembles.
\newblock {\em Advances in Neural Information Processing Systems}, 33, 2020.

\bibitem{islam2003constructive}
Md~M Islam, Xin Yao, and Kazuyuki Murase.
\newblock A constructive algorithm for training cooperative neural network
  ensembles.
\newblock {\em IEEE Transactions on Neural Networks}, 14(4):820--834, 2003.

\bibitem{russakovsky2015imagenet}
Olga Russakovsky, Jia Deng, Hao Su, Jonathan Krause, Sanjeev Satheesh, Sean Ma,
  Zhiheng Huang, Andrej Karpathy, Aditya Khosla, Michael Bernstein, et~al.
\newblock Imagenet large scale visual recognition challenge.
\newblock {\em International Journal of Computer Vision}, 115(3):211--252,
  2015.

\bibitem{liao2018defense}
Fangzhou Liao, Ming Liang, Yinpeng Dong, Tianyu Pang, Xiaolin Hu, and Jun Zhu.
\newblock Defense against adversarial attacks using high-level representation
  guided denoiser.
\newblock In {\em Proceedings of the IEEE Conference on Computer Vision and
  Pattern Recognition}, pages 1778--1787, 2018.

\bibitem{zhang2018deep}
Ying Zhang, Tao Xiang, Timothy~M Hospedales, and Huchuan Lu.
\newblock Deep mutual learning.
\newblock In {\em Proceedings of the IEEE Conference on Computer Vision and
  Pattern Recognition}, pages 4320--4328, 2018.

\bibitem{loshchilov2017decoupled}
Ilya Loshchilov and Frank Hutter.
\newblock Decoupled weight decay regularization.
\newblock In {\em International Conference on Learning Representations}, 2018.

\bibitem{clevert2015fast}
Djork-Arn{\'e} Clevert, Thomas Unterthiner, and Sepp Hochreiter.
\newblock Fast and accurate deep network learning by exponential linear units
  (elus).
\newblock {\em arXiv preprint arXiv:1511.07289}, 2015.

\bibitem{chen2020laplacian}
Xuejin Chen, Xiaotian Chen, Yiteng Zhang, Xueyang Fu, and Zheng-Jun Zha.
\newblock Laplacian pyramid neural network for dense continuous-value
  regression for complex scenes.
\newblock {\em IEEE Transactions on Neural Networks and Learning Systems},
  32(11):5034--5046, 2020.

\bibitem{zhang2020densely}
Jinqing Zhang, Haosong Yue, Xingming Wu, Weihai Chen, and Changyun Wen.
\newblock Densely connecting depth maps for monocular depth estimation.
\newblock In {\em European Conference on Computer Vision}, pages 149--165.
  Springer, 2020.

\bibitem{huynh2020guiding}
Lam Huynh, Phong Nguyen-Ha, Jiri Matas, Esa Rahtu, and Janne Heikkil{\"a}.
\newblock Guiding monocular depth estimation using depth-attention volume.
\newblock In {\em European Conference on Computer Vision}, pages 581--597.
  Springer, 2020.

\bibitem{Long_2021_ICCV}
Xiaoxiao Long, Cheng Lin, Lingjie Liu, Wei Li, Christian Theobalt, Ruigang
  Yang, and Wenping Wang.
\newblock Adaptive surface normal constraint for depth estimation.
\newblock In {\em Proceedings of the IEEE International Conference on Computer
  Vision}, pages 12849--12858, October 2021.

\bibitem{paszke2017automatic}
Adam Paszke, Sam Gross, Soumith Chintala, Gregory Chanan, Edward Yang, Zachary
  DeVito, Zeming Lin, Alban Desmaison, Luca Antiga, and Adam Lerer.
\newblock Automatic differentiation in pytorch.
\newblock In {\em Advances in Neural Information Processing Systems Workshop
  Autodiff}, 2017.

\end{thebibliography}
}

\end{document}